\documentclass{article}

\PassOptionsToPackage{numbers}{natbib}
\usepackage[final]{neurips_2024}

\usepackage[utf8]{inputenc} % allow utf-8 input
\usepackage[T1]{fontenc}    % use 8-bit T1 fonts
\usepackage{hyperref}       % hyperlinks
\usepackage{url}            % simple URL typesetting
\usepackage{booktabs}       % professional-quality tables
\usepackage{amsfonts}       % blackboard math symbols
\usepackage{nicefrac}       % compact symbols for 1/2, etc.
\usepackage{microtype}      % microtypography
\usepackage{xcolor}         % colors

\usepackage{array}
\usepackage{booktabs}
\usepackage{tabularx}
\usepackage{longtable}
\usepackage{multirow}

\usepackage{caption}
\usepackage{subcaption}

\usepackage{amsmath}
\usepackage{amssymb}
\usepackage{amsthm}

\usepackage{hyperref}

\usepackage{xcolor}

\usepackage{tcolorbox}
\usepackage[ruled]{algorithm2e} % http://www.ctan.org/pkg/algorithm2e
\usepackage{xcolor} % http://www.ctan.org/tex-archive/macros/latex/contrib/xcolor
\usepackage{tikz}
\usetikzlibrary{fit,calc}
\newcommand*{\tikzmk}[1]{\tikz[remember picture,overlay,] \node (#1) {};\ignorespaces}
\newcommand{\boxit}[1]{\tikz[remember picture,overlay]{\node[yshift=3pt,fill=#1,opacity=.25,fit={(A)($(B)+(.95\linewidth,.8\baselineskip)$)}] {};}\ignorespaces}
\colorlet{mypink}{red!40}
\colorlet{myblue}{cyan!60}
\colorlet{mygreen}{green!50} % Adding new color 'mygreen'

\definecolor{codegreen}{rgb}{0,0.6,0}
\definecolor{codegray}{rgb}{0.5,0.5,0.5}
\definecolor{codepurple}{rgb}{0.58,0,0.82}
\definecolor{backcolour}{rgb}{0.95,0.95,0.92}
\usepackage{listings}
\usepackage{wrapfig}
\usepackage{enumitem}
\usepackage{hyperref}
\usepackage{amsthm}
\newcommand{\Dist}[1]{\mathbf{\Delta}(#1)}

\usepackage{amsmath}
\usepackage{amssymb}
\usepackage{mathtools}
\usepackage{environ}
\usepackage[capitalize,noabbrev]{cleveref}
\theoremstyle{plain}
\newtheorem{theorem}{Theorem}[section]

\theoremstyle{definition}
\newtheorem{definition}[theorem]{Definition}

\theoremstyle{remark}

\newcommand{\method}{\textsc{FactorSim}\xspace}

\title{\method: Generative Simulation\\ via Factorized Representation}

\author{%
Fan-Yun Sun\\Stanford University \And 
S. I. Harini\\Stanford University \And 
Angela Yi\\Stanford University \And Yihan Zhou\\Stanford University \AND
Alex Zook\\Nvidia  \And Jonathan Tremblay\\Nvidia \And 
Logan Cross\\Stanford University \AND
Jiajun Wu\\Stanford University \And 
Nick Haber\\Stanford University 
}

\begin{document}

\maketitle

\begin{abstract}
% \sun{ Do the main claims made in the abstract and introduction accurately reflect the paper's contributions and scope?}

% Unsupervised generation of high-quality multi-view-consistent images and 3D shapes using only collections of single-view 2D photographs has been a long-standing challenge. 
% Existing 3D GANs are either compute-intensive or make approximations that are not 3D-consistent; the former limits quality and resolution of the generated images and the latter adversely affects multi-view consistency and shape quality. 
% In this work, we improve the computational efficiency and image quality of 3D GANs without overly relying on these approximations. For this purpose, we introduce an expressive hybrid explicit-implicit network architecture that, together with other design choices, synthesizes not only high-resolution multi-view-consistent images in real time but also produces high-quality 3D geometry. 
% By decoupling feature generation and neural rendering, our framework is able to leverage state-of-the-art 2D CNN generators, such as StyleGAN2, and inherit their efficiency and expressiveness. 
% We demonstrate state-of-the-art 3D-aware synthesis with FFHQ and AFHQ Cats, among other experiments. 

% Training intelligent agents, such as robots, is a non trivial process, 
% it requires carefully implementing simulations, 
% defining the learning task and its interactions, 
% wrapping the created simulation into a training framework, 
% \textit{etc.}
%%% new version
%Historically, people rely on zero-shot transfer 

Generating simulations to train intelligent agents in game-playing and robotics from natural language input, 
% \textit{e.g.}, 
from user input or task documentation, remains an open-ended challenge. 
Existing approaches focus on parts of this
challenge, such as generating reward functions or task hyperparameters.
%generating the task hyperparameters, 
%or populating the environment with assets while omitting key elements of the simulation logic.
% , game mechanics for example. 
% Unlike previous work, we aim to generate full simulations; 
%However, existing approaches are unable to do \todo{}.
Unlike previous work, we introduce \method that generates full simulations in code from language input that can be used to train agents. 
Exploiting the structural modularity specific to coded simulations, we propose to use a \textbf{factored} partially observable Markov decision process representation that allows us to reduce context dependence during each step of the generation.
%leverage that 
% coded
%simulations can be
% modeled as factored Partially Observable Markov Decision Processes (POMDPs). 
%modeled as factored partially observable Markov decision processes which allow us 
%to explore both this factorized representation and the model-view-controller paradigm. 
% In greater detail, 
%\method first decomposes generation into a series of steps with minimal necessary context given to the language model at each step, reducing the reasoning complexity of each step. 
For evaluation, we introduce a \textit{generative simulation} benchmark that assesses the generated simulation code's accuracy and effectiveness in facilitating zero-shot transfers in reinforcement learning settings.
% use a reinforcement learning benchmark and derive language descriptions of simulations from its documentation. 
% For evaluation, we use a reinforcement learning benchmark and derive language descriptions of simulations from its documentation. 
% We demonstrate that \method outperforms many competitive methods in generating full game simulation code from scratch both qualitatively and quantitatively,
% and achieves superior zero-shot transfer results after training on the generated
% simulations.
% We demonstrate that \method outperforms many competitive methods on these tasks and more, 
% both qualitatively and quantitatively.
We show that \method outperforms existing methods in generating simulations regarding prompt alignment (\textit{i.e.}, accuracy), zero-shot transfer abilities, and human evaluation. We also demonstrate its effectiveness in generating robotic tasks.
% and achieves superior zero-shot transfer results after training on the generated
% simulations.
\stepcounter{footnote}
\footnotetext{Work done while Harini S I was an intern at Stanford.}
\stepcounter{footnote}
\footnotetext{Correspondence to sunfanyun@cs.stanford.edu}
\stepcounter{footnote}
\footnotetext{Project website: \href{https://cs.stanford.edu/~sunfanyun/factorsim/}{https://cs.stanford.edu/~sunfanyun/factorsim/}}

\end{abstract}

%\vspace{-3mm}
\section{Introduction}

Simulations hold significant potential for training agents to perform real-world tasks where data collection is costly, dangerous, or infringes on individual privacy. 
% These tasks include robots folding laundry or navigating a warehouse, and agents playing games like Pong, either as players or non-player character (NPC) adversaries.
A major bottleneck in harnessing the potential of simulations at scale for agent training is the cost of designing and developing them, especially when we need a distribution of simulations that meet detailed design specifications to train more generalized policies. 
%In addition, training generalized policies requires generating a distribution of related simulation environments.
% Another crucial challenge is creating a diverse distribution of simulations to train reinforcement learning (RL) agents across a range of tasks, which is essential for developing generalized policies. 
% We aim to alleviate this simulation authoring bottleneck with techniques to generate robust simulations for agent training that adhere to detailed design specifications.
%Creating these simulations can be demanding for expert human users, which limits their scalability.
In this paper, we aim to generate coded simulations given text specifications. Code provides a natural interface for users to inspect, modify, and debug the simulation. It also allows us to craft diverse environments for Reinforcement Learning (RL) purposes.
% Since coding these simulations can be demanding for expert human users, which leads to limiting scalability (there are so many experts in the world), 
% we thus aim to automatically generate these distribution of simulations from text specifications.

% Large language models (LLMs) have emerged as a promising way for generating simulations specified as code.
% LLMs take user input as a text prompt and output the code for a simulation environment that is used for downstream tasks like agent training.
% While LLMs have shown promise for generating simple simulations~\cite{zala2024EnvGen}\zook{other ref?}, they struggle to generate a full simulation in a single step when provided detailed instructions~\cite{wang2023gensim,wang2023robogen}.
% Simulations are typically designed as a set of interacting components, thus it is natural to decompose simulation construction into generating these interacting pieces.
% Unfortunately, simply generating the simulation in a series of steps requires a large input context in the form of the code for the simulation so far.
% This is undesirable given the context length limitations of LLMs and their weaknesses in fully leveraging all information provided as input context~\cite{liu2024lost}.
% We instead propose to decompose the simulation into a set of interacting components and task the LLM with generating each of these components when provided the minimal context needed for their definition.

Generating full simulations in code to train agents from a text prompt is an under-explored challenge. 
% The process of generation simulations to train agents from text prompt is a somewhat new task that is holistically under explored. 
Previous works focus on parts of this challenge, 
including reward function design~\cite{ma2023eureka}, hyperparameter tuning~\cite{mandlekar2023mimicgen}, and task configuration while relying on an existing simulator~\cite{wang2023gensim}.
These methods use large language models (LLMs) to generate the components of simulations specified as code.
% The methods share in common the usage of Large language models (LLMs) has they have emerged as a promising way for generating simulations specified as code.
%The LLM takes a prompt of text describing a specific simulation and then generates the simulation code or components.
%A natural extension of this pipeline is allowing the LLM to self-debug and iterate on the task by using a provided compiler and a prior code base~\cite{chen2024teaching}.
% LLMs take user input as a text prompt and output the code for a simulation environment that is used for downstream tasks like agent training.
% While LLMs have shown promise for generating simple simulations~\cite{zala2024EnvGen}\zook{other ref?}, they struggle to generate a full simulation in a single step when provided detailed instructions~\cite{wang2023gensim,wang2023robogen}.
However, when faced with large and detailed contexts, LLMs often generate simulations that ignore or fail to adhere to parts of the input prompt~\cite{liu2024lost}. This issue is not solely due to the limitations of existing LLMs but also suggests that some form of decomposition is always critical as we scale up the number of components in simulations. We ask the question: can we exploit the inherent structure (e.g., having a game loop that handles agent actions, updates internal game states accordingly, and displays the game states to the users through a rendering process) of coded simulations to generate them better?

% JOHN: I feel this argument is a little all over the place and does not move the narrative correctly. I feel like we would need to back suck argument better in the paper. 
% Existing work uses methods like chain-of-thought (CoT) prompting to have the LLM break down the generation task into a series of reasoning steps or sub-tasks~\cite{wei2022chain,yao2022react,shinn2023reflexion}.
% These techniques often fail at complex code-generation or editing instructions as the CoT structure does not isolate the context needed for different parts of the code, leading to failure cases including hallucinating non-existent functions, opting out of generating all the code, or producing code with compile-time or runtime errors.

% John folded this argument into the one above. 
% Simulations are typically designed as a set of interacting components, thus it is natural to decompose simulation construction into generating these interacting pieces.
% Unfortunately, simply generating the simulation in a series of steps requires a large input context in the form of the code for the simulation so far.
% This is undesirable given the context length limitations of LLMs and their weaknesses in fully leveraging all information provided as input context~\cite{liu2024lost}.
% We instead propose to decompose the simulation into a set of interacting components and task the LLM with generating each of these components when provided the minimal context needed for their definition.

\begin{figure}
    \centering
    \includegraphics[width=\textwidth]{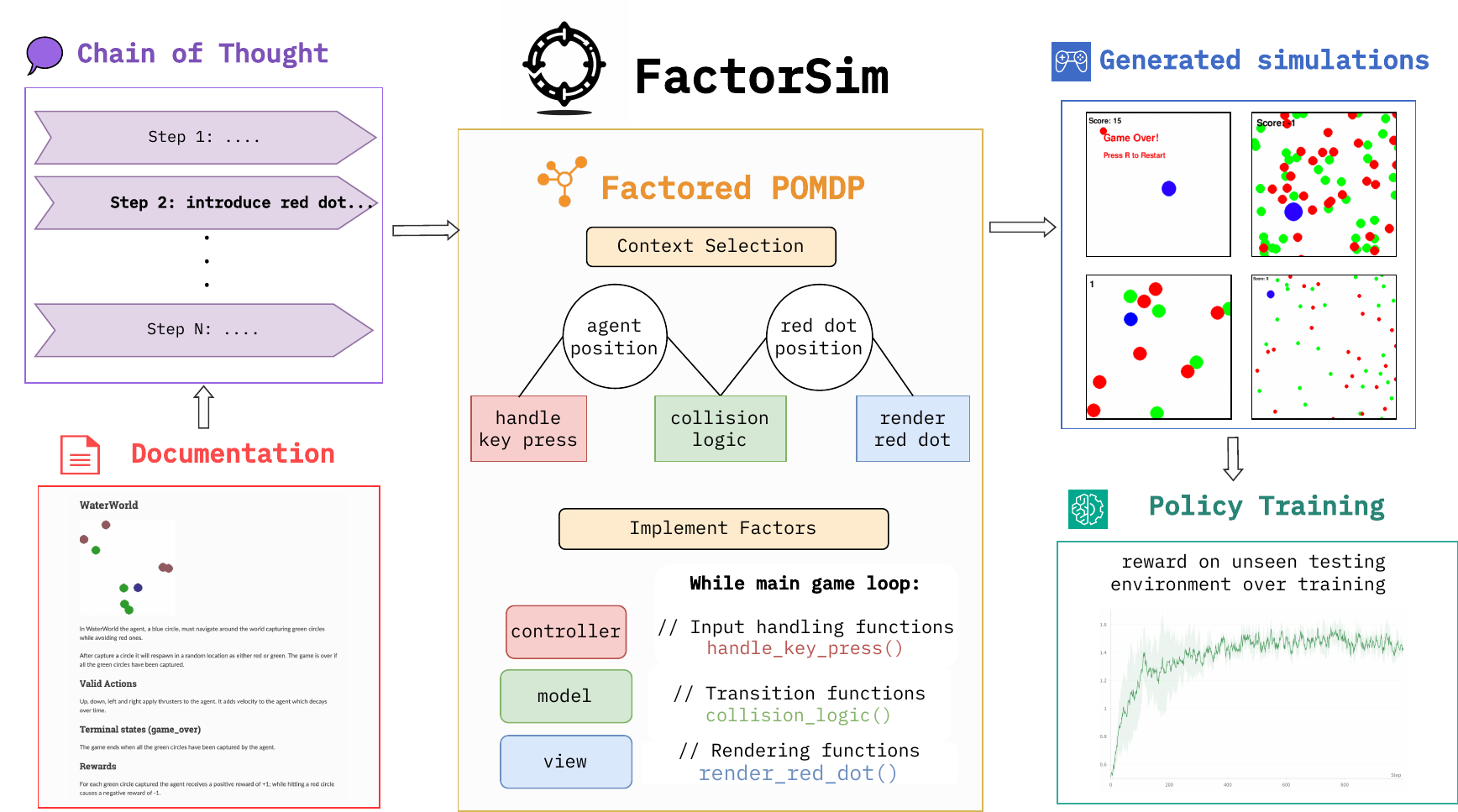}
    \caption{
    Overview of \method{}. \method takes language documentation as input, uses Chain-of-Thought to derive a series of steps to be implemented, adopts a Factored POMDP representation to facilitate efficient context selection during each generation step, trains agents on the generated simulations, and tests the resulting policy on previously unseen RL environments.
    %Given a language description of a simulation, our proposed method first factorizes the prompt into POMDPs components, such as 
    %state, state transition, reward function, etc. (top right).
    %We represent this process through color encoding, where it is shown that the text highlighted in a specific color is used in specific nodes for its generation. 
    %Given the graph composition, the method composes specific prompts for a LLM to generate code for the simulation (bottom left).
   %At the end of this process, we end up with a simulation in adherence to the initial prompt (shown on the bottom right). 
    %When the generation process is repeated, its stochastic nature gives us different sample of the expected simulation which is useful for training general AI agent on a distribution of tasks. 
    %\john{some notes: code generation is additive (alex) maybe having this shown instead of code generation maybe symbol. It is random because of LLM nature, not design (alex). might want to highlight RL training (sun). might want to highlight multiple generation env (sun). }
    }
    \label{fig:overview}
    %\vspace{-12mm}
\end{figure}

% Our insight is that by viewing the generation process through the lens of a factorization of a probability distribution we can identify the minimal context required for addressing each sub-problem.
% We propose to model simulations as Partially Observable Markov Decision Processes (POMDPs) and leverage a factorized representation of their structure to define a decomposition of the LLM input into a set of smaller prompts that have reduced context dependencies.
We propose \method, a framework that takes an arbitrary language specification as input and outputs a full simulation that can be used to train RL agents. The key idea of \method is to decompose the input prompt into a series of steps and then use a factored Partially Observable Markov Decision Process (POMDP) representation to reduce the context needed for each generation step. To realize \method, we use the \textit{model-view-controller} software design pattern to structure the generation process. 
Consider generating a coded simulation of WaterWorld; see Figure~\ref{fig:overview}. The game consists of an agent (blue circle) traveling in a 2d world, capturing food (green circle) while avoiding enemies (red circle). 
Our method first decomposes the game description into multiple steps to be implemented. For example, a step instruction could be ``Introduce red dot enemies that can be controlled with arrow keys. Give the player a -1 reward when the agent collides with an enemy''. 
We first select the context needed for this functionality to be implemented, \textit{e.g.}, positions of existing agents. 
Subsequently, \method generates (at most) three functions: one to handle player input (i.e., \textit{handle\_key\_press}, the controller component), one to implement the collision logic (i.e., \textit{collision\_logic}, the model component), and one to update the rendering function (i.e., \textit{render\_red\_dot}, the view component). 
Limiting the context during each step of the simulation generation process allows \method to focus on the task at hand while avoiding hallucinating non-existent functions or modifying code not meant to be changed.
%reasoning about the game logic, and updating simulation's visuals. 
%Subsequently, to reduce the context of each generation step, we propose to model simulations as . 
%In other words, 
% process we leverage that for each module there exists a minimal context window of the simulation.   
% This smaller context window also allow use to leverage 
 %, this allows for a greater decomposition and it also helps reducing the context for subsequent modules. 

To evaluate the task of full simulation generation, we propose a new \textit{Generative Simulation}\footnote{We adopt this term from \cite{xian2023towards} to refer to automated simulation generation to train agents within.} benchmark with accompanying success metrics.
One set of success metrics is the pass rate in automated system tests. Commonly used in game development, these system tests programmatically assess whether the behavior of the generated simulation adheres to the specifications given in the input prompt. 
%This evaluates whether the generated simulation aligns with the specification in the prompt. 
The second success metric assesses the value of the generated simulations for transfer learning in an RL setting. 
This evaluates how well agents trained on a set of generated simulations can generalize to held-out environments that satisfy the design specifications provided in prompts.
Generalization to unseen environments is crucial for many applications, including transferring robotics policies learned in simulation to the real world.
% JOHN i would remove this last sentence, since it is not a contribution. 
This benchmark consists of 8 RL environments with varying levels of difficulty. In addition to evaluating our method on the benchmark we introduced, we further assess \method's ability to generate robotic tasks on the dataset published by GenSim~\cite{wang2023gensim}. We demonstrate the value of our method, \method, on both the benchmark task suite and GenSim's dataset, showing performance superior to baseline alternatives.
In summary, our contributions are three-fold. First, we propose \method, a framework for generating coded simulation with a factor graph of a POMDP as a principled way to reduce context dependence. Second, we introduce a new generative simulation benchmark by adapting an existing RL benchmark~\cite{tasfi2016PLE}, and demonstrate \method's superior results against baselines in terms of code correctness (i.e., prompt alignment), ability to facilitate zero-shot generalization and human evaluation of the simulations.
Third, we demonstrate that \method can be applied to generating simulation tasks for robotics, outperforming existing approaches.

%\zook{perhaps an evaluation criteria we can employ is how much context was needed by the LLM}
% why is this valuable?
% - we show how to improve control over generated simulations when providing detailed simulation design instructions
% - our method works generally for any POMDPs
% - we reduce the amount of context needed to prompt the LLM, which is generally useful

%\vspace{-2mm}
\section{Related Work}

We aim to generate simulations for training agents to generalize to previously unseen environments.
Recent work has investigated this in the context of learned neural world models and LLM-generated code for simulations.

World models simulate the dynamics of a given environment and use this as a proxy environment for agent training, rather than interacting with a ground truth simulator~\cite{ha2018worldmodel}.
Several approaches have demonstrated the value of learning world models as part of general algorithms that can learn to play a variety of games (AlphaZero~\cite{silver2018alphazero}, Muesli~\cite{hessel2021muesli}, and DreamerV3~\cite{hafner2023dreamverv3}).
% They have been used to train agents to play a variety of games, including AlphaZero~\cite{silver2018alphazero}, Muesli~\cite{hessel2021muesli}, and DreamerV3~\cite{hafner2023dreamverv3}.
Other efforts use a large set of offline data to learn a world model that is subsequently used for agent training, including for autonomous vehicle driving (GAIA-1~\cite{hu2023gaia1}), robotic manipulation (UniSim~\cite{yang2024unisim}), and 2D platformer games (Genie~\cite{bruce2024genie}).
% GAIA-1~\cite{hu2023gaia1} learns a world model for autonomous vehicle simulation from vehicle driving data and Genie~\cite{bruce2024genie} learns a world model capable of generating a variety of platform games by learning a latent action model from online videos without action labels.
% UniSim~\cite{yang2024unisim} learns a world model from a combination of human and robot demonstrations with image and video data, then demonstrates successful reinforcement learning policy training in these learned simulations.
% SIMA~\cite{raad2024sima} learns a general 3D game playing policy from human demonstrations where the world dynamics are implicitly captured in the learned policy due to it's generality across environments.
% We choose to focus on world models specified as code instead of learned world models.
We generate world models as code as they are more interpretable, modular, and easily modified or extended by humans---key advantages we believe are important for their use in authoring large-scale or complex simulations.

LLMs have generated many parts of simulations for game playing and robotics.
%, including the specification of the environment, task structure, and reward functions for agent training.
In (RL) games, LLMs have been used to generate game levels~\cite{todd2023level, sudhakaran2023mariogpt}, to choose parameters for an existing simulator~\cite{zala2024EnvGen}, and to assist humans in creating full games~\cite{anjum2024inksplotch}.
% Unlike these approaches, we focus on the task of generating all the code needed to specify a simulator (game) needed to train a reinforcement learning agent without requiring human intervention during the generation process.
%
In robotics, LLMs have been used to generate reward functions, task specifications, and specific components like scene configurations within robotics tasks. 
%and choose learning algorithms for agents trained on these tasks.
Many works such as RoboGen~\cite{wang2023robogen}, Holodeck~\cite{yang2023holodeck}, and Gen2Sim~\cite{katara2023gen2sim} build on top of existing simulators and use a series of prompts to generate interactable 3D environments to train agents. 
%~\cite{wang2023robogen,katara2023gen2sim,yang2023holodeck}. 
%RoboGen~\cite{wang2023robogen} uses a series of prompts to subdivide the process of generating 3D scenes and tasks to train agents. 
% for agents to acquire skills within. %, and then has an LLM choose how to train an agent appropriate to the generated tasks.
GenSim~\cite{wang2023gensim} starts from a human task library and iteratively generates and tests new tasks to generate robotic manipulation tasks.
%PhyScene~\cite{yang2024physcene} uses a diffusion model to generate room layouts for physically interactable scenes. % not about LLMs per se
Other efforts have focused on generating reward functions for tasks~\cite{ma2023eureka,kwon2023llmreward,ma2024dreureka}. 
Eureka~\cite{ma2023eureka} uses feedback from agent training to refine reward function specification. 
% , and DeLF~\cite{afshar2024delf} generates the state and action spaces of an MDP
% Eureka~\cite{ma2023eureka} demonstrated LLMs can use information about a reinforcement learning agent on robotics tasks to improve the specification of the task reward function defined in code.
% DeLF~\cite{afshar2024delf} instead generates state and action spaces for an MDP, but does not address reward specification.
Our approach is able to generate both the simulator dynamics and reward functions and can be applied to both robotics and games.

% While these approaches demonstrate the potential for LLMs in robotics task generation, they do not leverage knowledge of the underlying MDP structure to define the task or rewards.
%
% Other efforts have studied generating parts of the MDP defining an environment.
% DeLF~\cite{afshar2024delf} generates state and action spaces for an MDP, but does not address reward specification.
% Many efforts use an LLM to generate reward functions for tasks (Eureka~\cite{ma2023eureka}, RoboGen~\cite{wang2023robogen}), but leave the underlying simulator fixed.
% Our approach jointly specifies the simulator dynamics and rewards: naturally complementing approaches that improve generation with feedback from execution.

% CoT &co
As noted above, LLMs can struggle to handle complex tasks: this has prompted research into different ways to structure LLM reasoning.
% Early work demonstrated that transformer models can use an external scratchpad to improve performance over multiple inference iterations~\cite{nye2021scratchpad}.
Chain-of-Thought (CoT) prompting demonstrated LLM performance can be substantially boosted by prompting the LLM to break a single task into multiple steps with either few-shot examples~\cite{wei2022chain} or zero-shot~\cite{kojima2022zeroshotcot}.
Subsequent work has developed a variety of techniques to improve LLM reasoning through multi-step reasoning prompts: checking for consistency among multiple reasoning paths~\cite{wang2023selfconsistency}, interleaving reasoning and tool use (ReACT~\cite{yao2022react}), using tree data structures to guide the LLM reasoning process (Tree-of-Thought~\cite{yao2024tree}), or formulating reasoning as a tree search process~\cite{hao2023rap,zhou2023lats}.
Approaches for general code generation include decomposing the task into functions to subsequently generate (Parsel~\cite{zelikman2023parsel}), generating code to reach a series of intermediate execution states (ExeDec~\cite{shi2023exedec}), and using a multi-agent framework to generate, test, and refine code (AgentCoder~\cite{huang2023agentcoder}).
% and Language Agent Tree Search~\cite{zhou2023lats}).
Other efforts optimize the prompts for given tasks, using evolutionary search (EvoPrompt~\cite{guo2024connecting}) or defining generalized declarative programming frameworks with modular optimization algorithms~\cite{khattab2024dspy}.
% Our approach uniquely leverages the MDP structure of the task to provide a decomposition that minimizes input context needed for different reasoning steps and can be readily combined with techniques that optimize the discrete tokens used in input prompts.
Our approach generates code by leveraging a factorized representation specific to simulations to reduce the input context needed for different reasoning steps; it can be used in conjunction with approaches for general code generation, such as generating tests as a form of self verification.%incorporating feedback.

\section{\method{}: Generating Simulations via Factorized Representation}

A simulation is a structured system of modules connected by events and responses. Our framework, \method, generates code using LLMs by exploiting this structure to construct a simulation progressively. Our key insight is that, by generating a simulation step-by-step while \textbf{only selecting the relevant context information needed for each step}, we can effectively reduce the reasoning capacity needed for each step, leading to simulations that adhere more closely to the input requirements.
%Exploiting that structure allows code generation to proceed in steps that each stay within the reasoning capacity of existing LLMs, leading to simulations that adhere more closely to the input requirements.

In this section, we describe our method for generating Turing-computable simulations.
First, we describe simulations that can be modeled as a Partially Observable Markov Decision Process (POMDP).
Second, we use Chain-of-Thought (CoT) to decompose an input prompt describing the desired full simulation into a series of prompts describing different components to be implemented.
Third, we introduce a factorized POMDP representation that exploits the inherent modularity of coded simulations.
%to provide the output targets for the series of prompts.
%Third, we further decompose the POMDP state transition function to mitigate common weaknesses in this part of tightly coupled simulation code .
%Finally, we show how the factor graph representation of the POMDP defines the minimal context needed to generate each step, reducing the input context needed in each generation step.
%Subsequently, we introduce our framework, \method{}, that leverages the representation of POMDPs to structure code generation. 
% Algorithm~\ref {alg:factor} shows at a high level the proposed generation process. 
%\method{} exploits the inherent modularity of simulations by generating individual modules, each defined by \textit{the minimal context needed} required for its creation.
% For example, when asked to generate code for a state transition function, it only uses information about the state and the query, but not how the states are rendered.
% The following sections introduce the formalization of this process and explain the intuition behind it. 
% \john{needs a little bit more about the later subsection on division of the f and g. and other things}
Refer to Algorithm 1 and Figure~\label{fig:method_illustrative_overview} for an overview of \method{} alongside an illustrative example.

%\captionsetup[algorithm]{font=small}

\begin{algorithm}
\small
\label{alg:factor}
\caption{\method{}}
\DontPrintSemicolon % Suppresses semicolon at the end of each line
\KwIn{$Q_{\text{text}}$, a natural language description of the simulation, and an LLM}
\KwOut{a turing-computable simulation represented as a POMDP $\mathcal{M}^\prime = \langle S, A, O, T, \Omega, R \rangle$}
\;
Initialize a Factored POMDP $\mathcal{M}_1 \gets \langle S_1, A, \emptyset, T_1, \emptyset, R_1 \rangle$ where \; 
- $S_1 := \{ s_{\text{score}} \}$\;
- $A$ is the set of all keyboard inputs\;
- $T_1$ is an identity function, i.e., $T_1(s' \mid s, a) = \mathbf{1}[s' = s]$\;
- $R_1(s, a, s') := s'_{\text{score}} - s_{\text{score}}$\;\;
// Chain of Thought \;

% Decompose $Q_{\text{text}}$ into a list of modules $q_{\text{text}}$\;
Derive a step-by-step plan $(q_1,\ldots,q_k) \sim p(q_1,\ldots,q_k \mid Q_{\text{text}})$ \hfill Eq.~\eqref{eq:chain_of_thought}{\parfillskip0pt\par}\;
\For(){\text{each step, or module} $q_{k}$}{
    % \textbf{State space update} $p(S^* | \mathcal{M}, q_{\text{text}})$\;
    %\textbf{State space update} $S \gets S  p(S | \mathcal{M}, q_{\text{text}})$\;
    \textbf{State space update \& context selection} $ p(S_{k+1}, S\left[Z_{k}\right] | S_k, q_k)$ \hfill  Eq.~\eqref{eq:update_state_space},\eqref{eq:identify_scope}{\parfillskip0pt\par}\;
    \tikzmk{A} 
    // Controller component update\;
    \textbf{Action-dependent state transition model update}: $p(T^{(a)}_{k+1} | S\left[Z_{k}\right], A, q_k)$\;
    \tikzmk{B}
    \boxit{mypink}
    \tikzmk{A} 
    // Model component update\;
    \textbf{Action-independent state transition model update}: $p(T^{(s)}_{k+1} | T[Z_{k}], S\left[Z_{k}\right], q_k)$ \;
    %\textbf{State transition model update}: \;%$\{g_i\} \gets p(T^\prime | T, \{f_i\}, S^\prime, q_k)$ \;
    %\textbf{Reward model update}: \;%$R^\prime \gets p(R | R, S^\prime, q_k)$\; 
    \tikzmk{B} \boxit{mygreen}
   \tikzmk{A} 
    // View component update\;
   \textbf{Observation model update}: $p(\Omega_{k+1} | S\left[Z_{k}\right], q_k)$  \hfill Eq.~\eqref{eq:rendering}{\parfillskip0pt\par}\; 
   \tikzmk{B}
   \boxit{myblue}
    $\mathcal{M}_{k+1} = \langle S_{k+1}, A, O_{k+1}, T_{k+1}, \Omega_{k+1}, R_1 \rangle$ where $O_{k+1}$ is the new observation space defined by $S_{k+1}$ and $\Omega_{k+1}$, and $T_{k+1}(s^\prime \mid s, a) =  T^{(s)}_{k+1}(s^\prime \mid s) \cdot T^{(a)}_{k+1}(s \mid s,a)$. \;
}
Return the final simulation $\mathcal{M}^\prime \gets \mathcal{M}_{k+1}$\;
\end{algorithm}
%\nick{I think this is fairly clear. Small thing...presumably the unprimed variables need to be replaced with the primed variables in the for loop}

\subsection{Modeling Simulation as POMDP}
A Partially Observable Markov Decision Process (POMDP) is used to represent a coded simulation. Formally a POMDP is represented as a tuple 
$\mathcal{M} = \langle S, A, O, T, \Omega, R \rangle$ 
where 
$S$ is a set of states, 
$A$ is a set of actions, 
$O$ is a set of observations,  
%$T: \mathcal{P}_{S \times A, S}$ is a transition probability distribution, 
$T: S \times A \rightarrow \Dist{S}$ is a transition probability distribution, 
%$\Omega{}: \mathcal{P}_{S,O}$ is an observation distribution, 
$\Omega{}: S \rightarrow \Dist{O}$ is an observation function, and
%$R: \mathcal{P}_{S \times A, [0,1]}$ is the reward distribution, 
$R: S \times A \times S^\prime \rightarrow \mathbb{R}$ is the reward model~\footnote{We omit the discount factor $\gamma$ and the initial state distribution $\pi$ in the formulation for brevity. In our experiments, $\pi$ is generated alongside the states $S$.}. 
%We adopt a simplified observation model as in~\cite{dennis2020emergent} where the partial observability function is independent of actions.
%, and the reward model is independent of the next state.

% \subsection{\method{}}

% \subsection{\method{}}

%%%%%%%%%%%%%%%%%%%%%%%%%%
% What is the problem, mathematically?
%%%%%%%%%%%%%%%%%%%%%%%%%%
% When looking at the structure of the prompt, $Q_{\text{text}}$, 
We aim to generate a simulation from a prompt $Q_{\text{text}}$. In this paper, we are particularly interested in the case where $Q_{\text{text}}$ comprises detailed design specifications such that the resulting simulation could be used to train agents, though our method applies to any prompt for defining a simulation. In our experiments, $Q_{\text{text}}$ is a paragraph of text around 10 sentences specifying this simulation.
%As noted above, using LLMs to directly sample an entire POMDP $\mathcal{M}$ conditioned on text $Q_{\text{text}}$ can result in simulations that fail to adhere closely to the provided prompt.
% When sampling a simulation in a freeform way, $\mathcal{M} \sim p(\mathcal{M} | Q_{\text{text}})$, which could be done through an LLM.
% Without any guarding rails, the output could be quite chaotic and inaccurate. 
% In the context of generating simulations for training autonomous agents, it is crucial that the simulation environment adheres to the rules and conditions specified.
%Without any guarding rails, the output could be quite chaotic and unstructured, such as from all the possible simulations that exist, 
%there are a multitude that are usuable.
% For example, in the game pong, when the ball hits a pallet it bounces, when the ball passes the opponent pallet you get 
% a point, \textit{etc.} 
% This process could generate a simulation whereas it confuses opponent pallet, your pallet, and the logic.
% And this could lead to the game deciding giving you a point when it hits any pallet. 
% Consider the game Pong: 
% The ball rebounds off a paddle upon contact and scores a point if it bypasses the opponent’s paddle. 
% A flawed simulation might mistakenly award a point each time the ball contacts any paddle, leading agents to learn incorrect strategies or behaviors.
% 
\vspace{-1mm}
\subsection{Chain of Thought}
% 1) from single prompt to plan
% 2) decompose POMDP
% 3) split out model-view-controller
% 4) leverage factor graph to reduce context
We first decompose the prompt $Q_{\text{text}}$ into a series of steps using Chain of Thought~\cite{wei2022chain}, each describing a module of the simulation to be implemented.
%%%%%%%%%%%%%%%%%%%%%%%%%%
% Why do we generate things progressively?
%%%%%%%%%%%%%%%%%%%%%%%%%%
% 1) from one prompt to planned sequence of prompts
% 
Following similar formulation as in~\cite{prystawski2024think}, this can be thought of as marginalizing over a step-by-step plan variable $(q_1,\ldots,q_k)$ using $N$ Monte Carlo samples:
\begin{align} \label{eq:chain_of_thought}
\hat{p}(\mathcal{M}' | Q_\text{text}) = \frac{1}{N}\sum^N_{i=1} p(\mathcal{M}' | q^{(i)}_1,\ldots,q^{(i)}_K), \;\;\;\text{where} \;(q^{(i)}_1,\ldots,q^{(i)}_K) \sim p(q_1,\ldots,q_K | Q_\textit{text}),
\end{align} 
$p$ is a probability estimation model (i.e., an LLM in our experiments), and $\mathcal{M}'$ is the resulting code that fully specifies a simulation. % \nick{What is this [] notation? Is this denoting a text sequence? Let's say a bit more explicitly what this sampling is? mathcal is kind of a funny choice for describing a single distribution}
In practice, we only produce a single plan $N=1$. 

Intuitively, this process breaks the prompt into sub-tasks.
After we sample such a plan of $K$ steps, we generate the simulation progressively.
Given an existing POMDP $\mathcal{M}$ and a natural language specification $q$, we update the POMDP to reflect the changes specified.
%\begin{align}
%p(\mathcal{M}' | [q^{(i)}_1,\ldots,q^{(i)}_k]) \approx p(\mathcal{M}' | \mathcal{M}_j, q^{(i)}_j) \hspace{2mm} j = 1 \ldots k
%\end{align}
\begin{align}
p(\mathcal{M}_{K+1} | q_1,\ldots,q_K) \approx \prod^K_{k=1}p(\mathcal{M}_{k+1} | \mathcal{M}_k, q_k)  
\end{align}
where $\mathcal{M}_{k+1}$ is the POMDP (simulation as code) after the $k$-th step is implemented, and $\mathcal{M}_{K+1}$ is the final simulation. % In our experiments, $Q_{\text{text}}$ can be a paragraph of more than 10 sentences comprising of specifications of this simulation, and an example for $q_k$ is shown in Figure~\ref{fig:running_example}.
While Chain-of-Thought prompting allows LLMs to avoid having to generate code for all simulation logic at once, 
the complexity of each step still grows with $k$ due to the expanding codebase. This task remains challenging because LLMs must comprehend the code and accurately identify where modifications are needed.
% 
% However, directly sampling from a large language model (LLM) to update a simulation according to the updates specified in step $q_k$ remains a significant challenge. As illustrated by the purple blocks in Figure~\ref{fig:running_example}, the context for this update, $M_k$, could easily consist of hundreds of lines of code, and LLMs must not only comprehend the code but also accurately determine where modifications are needed. 
Acknowledging the limited reasoning ability of LLMs, we ask: can we further decompose the $p(\mathcal{M}_{k+1} | \mathcal{M}_k, q_k)$ into simpler distributions to reduce the complexity of each prompt?
%\sun{clarify the notation more.} \nick{IS this meant to represent a factored decomposition? As in, you have these intermediate random variables $\mathcal{M}_k$ induced by successively increasing the complexity of the simulation? An example would be helpful for at least me, right now}
% This is analogous to papers that have LLMs come up with a step-by-step plan before execution~\cite{yao2022react}. 
% In the previous mathematical construction, we represent the process of adding to POMDP using the prime notation, \textit{e.g.}, $\mathcal{M}'$ is the results of processing a text query, $q_k$, from the previous simulation $\mathcal{M}$.
%This process is sequentially executed to generate the final simulation
%

%\sun{give examples?}\zook{perhaps we have a running pong example that we reference in these steps}.
% Note that the model still has to take the existing simulation as context. 
% In our experiments, we ablate this version of factorization with our method and show that our method generates simulations with better prompt alignment. %adhere to the prompt at a much higher rate.

%\begin{figure}
%    \centering
%    \includegraphics[width=\textwidth]{figures/fig2.pdf}
%    \caption{\sun{TODO}}
%    \label{fig:running_example}
%\end{figure}

\vspace{-1mm}

\begin{figure}[!t]
  \centering
  %\vspace{-10mm}
  \includegraphics[width=\linewidth]{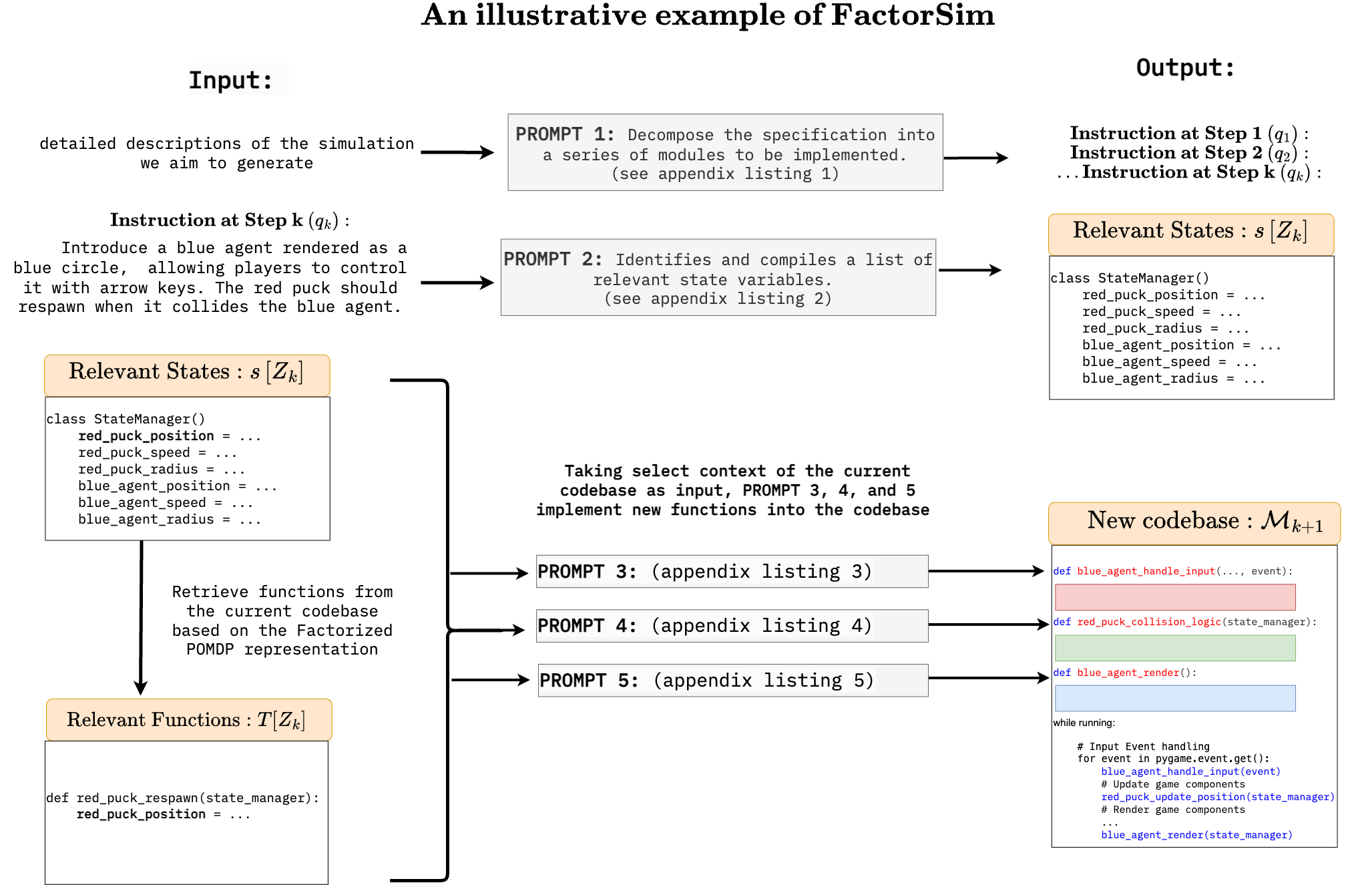}
  %\vspace{-3mm}
   \caption{An illustrative example of how the five main prompts in FactorSim correspond to our formulation in Algorithm 1. Note that the function \textit{red\_puck\_respawn} is retrieved as part of the context to Prompt 3, 4, and 5 because it modifies the state variable \textit{red\_puck\_position}, a state variable LLM identified as relevant in prompt 2.}
   \label{fig:method_illustrative_overview}
\end{figure}
\subsection{Decomposition by Factorized Representation}
%\sun{TODO: ground this subsection and the next to the example in the figure}
%%%%%%%%%%%%%%%%%%%%%%%%%%
% How do we exploit the structure of POMDP? 
% (a) model-view-controller paradigm
% (b) the intrinsic structure of a POMDP
%%%%%%%%%%%%%%%%%%%%%%%%%%
%To exploit the POMDP structure, we decompose the distribution as:
% To further decompose the POMDP structure, one could naively decompose the distribution as such:
Naively, we could further decompose a step of the generation into several steps, each focused on generating a different component of the POMDP:
\begin{align}
p(\mathcal{M}_{k+1} | \mathcal{M}_k, q_k) =  & p(S_{k+1} | \mathcal{M}_k, q_k) \cdot \label{eq:naive_state}\\
   & p(T_{k+1} | S_{k+1},  \mathcal{M}_k, q_k) \cdot \\
   & p(R_{k+1} | S_{k+1}, T_{k+1}, \mathcal{M}_k, q_k) \cdot \\
   & p(\Omega_{k+1} | S_{k+1}, T_{k+1}, R_{k+1}, \mathcal{M}_k, q_k) \label{eq:naive_rendering}
  % & p(\pi' | S', T', R', \Omega', \mathcal{M}, q_k) 
\end{align}
However, this still requires the LLMs to take the entire simulation ($\mathcal{M}_k$) as context, which could be over hundreds of lines of code in our experiments. Empirically, we observe that many failed generations can be attributed to LLMs attending to or modifying parts of the input context unrelated to the prompt.

% Here we describe our approach how to decompose POMDP generation in a way that reduces the amount of input context needed by each generation step.
%Here, we describe our approach for decomposing POMDP generation

%for the transition function \nick{Rephrase? Here you're defining what scope set means. What are the $\mathcal{T}_i$? I think giving some intuition for what these are (ideally with an example) would make it much easier to understand where this is going.}. 
%Our idea is to progressively generate a full simulation by treating each generation step as generating a factor. As such, during each step, we can only select specific components of the POMDP as context.
%(i.e. context state variables and context factors)to build our POMDP. 
%\sun{talk about how we select factors using the factorized states}
%By limiting the context needed in each step, we effectively reduce the complexity of the tasks, see Figure~\ref{fig:workflow} which compares generation to our proposed iterative process. 

To reduce the input context needed for each generation step, we propose to use a factored POMDP representation to remove the dependence on the full previous POMDP as context. For instance, given an existing simulation $M_k$ of red, green, and blue agents, to implement the $kth$-step instruction $q_k$: \texttt{respawn the red agent when it collides with the blue agent}, we only need context regarding the respawn logic of the red agent and the positions of the red and blue agents. Code regarding the green agent or the rendering logic would be unnecessary context.

To formalize our approach, we first introduce notation common to the literature~\cite{osband2014near,szita2009optimistic}.
% Following the definition in~\cite{osband2014near}. 
%In other words, factorization is the process of reducing a processing function into smaller, concise functions. \nick{What's a processing function?}
%This allows us to reason about the set of all possible functions that exists, whereas 
%a POMDP instance is now composed of a subset of that super set. \nick{I think this needs to be made more concrete, seems kind of tautological right now}
%Let's denote $Z \subseteq\{1,2, . ., n\}$ to define the scope set $\mathcal{T}[Z]:=\bigotimes_{i \in Z} \mathcal{T}_{i}$,
% $x\left[Z_{i}\right] = s\left[Z_{i}\right] \times A$, 
%for example, the action for movement is defined by using key arrows. \nick{Why does this expression depend on a single scope factor $Z_i$? Why is $A$ part of this for every $i$?}
%As such, we can always represent the factorized state and action space separately and rewrite Equation~\ref{eq:factor} to
%\begin{align}
%    T(s^\prime | s, a)=\prod_{i=1}^{m} T_{i}\left(s^\prime[i] \mid s\left[Z_{i}\right], a\right).
%\end{align}
%\nick{What do products mean here? These are actual functions, right, not distributions, so, these are for a decomposition of states into coordinates? But, reward is a single number, so this is a product? Or are these meant to represent distributions?}
%
Suppose we have a POMDP with a state space factored into $n$ state variables $S = S[1] \times \ldots S[n]$ and $Z$ is a subset of indices \( Z \subseteq \{1, 2, \ldots, n\} \), we define the scope set \(S[Z] := \bigotimes_{i \in Z} S[i] \) as the state space spanned by the subset of state variables. For example, if $Z = {1, 3, 4}$, then $S[Z]$ defines a state space defined by $S[1] \times S[3] \times S[4]$. We denote a state in the scoped state space $S[Z]$ as $s[Z]$. Below, let us formally define a factored POMDP.
%Another intuitive way is to think of 
% where \( n_k \in \mathbb{N} \) and \( n_k = t + s \) representing all factors \( t \) and  states \( s \) at step \( k \) respectively, 
%Let $\mathcal{T}_{S, A}$ define the space of all possible turing-computable~\footnote{A function that can be expressed with a computer algorithm (i.e. code).} transition functions that operate on the state space $S$ and action space $A$. We define a factored transition probability distribution and a factored reward function as follows:
\begin{definition}
A factored POMDP is a POMDP with both factored transition distribution and factored reward function. A transition probability distribution $T$ of a POMDP with discrete action space is factored over its state space $S = S_{1} \times \ldots S_{n}$ with scopes $Z_{1}, \ldots, Z_{m}$ if, for all $s \in \mathcal{S}, a \in A$ there exist some $\left\{T_{i} \right\}_{i=1}^{m}$ in the space of all possible transition distributions on the state space $S$ and action space $A$, such that,
\begin{align} \label{eq:factor}
      T(s | s, a)=\prod_{i=1}^{m} T_{i}\left(s[i] \mid s\left[Z_{i}\right], a\right).
\end{align}
A reward function $R$ of a POMDP is factored over $S = S_{1} \times \ldots S_{n}$ with scopes $Z_{1}, \ldots, Z_{l}$ if, for all $s \in \mathcal{S}, a \in A$ there exist some $\left\{R_{i} \right\}_{i=1}^{l}$ in the space of all possible reward functions on the state space $S$ and action space $A$, such that,
\begin{align}
    R(s, a)=\sum_{i=1}^{l} R_{i}\left( s\left[Z_{i}\right], a\right).
\end{align}
\end{definition}
%This process of factorizing also holds for reward distribution $\mathcal{R}$. 
%A factored POMDP is essentially a POMDP with both factored transition distribution and factored reward function. Note that the action space of the simulations we aim to generate is always discrete (i.e. fixed to a discrete set of \textit{keyboard inputs}).
% Borrowing from the literature of probabilistically graphical model, we use \textit{factor} to refer to functions that define the relationships between state variables.
%If we treat the reward model as part of the state transition function that updates a \textit{score} variable, 
A factored POMDP can be represented as a factor graph~\footnote{More precisely, a factor graph of a Dynamic Bayesian Network (DBN)~\cite{hansen2000dynamic,boutilier1996computing}.} with two types of nodes: \textit{state variables} (i.e., $S_i$) and \textit{factors} (i.e., $T_i$ or $R_i$), functions of (state) variables. Our idea is to \textbf{reduce context dependence by structuring the code using a factored POMDP representation} and treat each generation step as expanding a factored POMDP with new state variables and new \textit{factors}. During every step $q_k$, we first select a set of relevant state variable indices $Z_k$. Then, we select existing factors that have overlapping scope with the selected set of state variables as context, which we denote as $T[Z_{k}]$ and $R[Z_{k}]$. That is, we can reduce the dependence on the previous simulation $M_k$ and rewrite Equation~\ref{eq:naive_state}-\ref{eq:naive_rendering} to the following:
\begin{align}
  p(\mathcal{M}_{k+1} | \mathcal{M}_k, q_k) \approx  & p(S_{k+1}| S_k, q_k) \cdot  & \text{update state space} \label{eq:update_state_space}\\
   &  p( S\left[Z_{k}\right] | S_{k+1}, q_k) \cdot  & \text{identify relevant state variables} \label{eq:identify_scope} \\ 
   & p(T_{k+1} | T[Z_{k}], S\left[Z_{k}\right], A, q_k) \cdot  & \text{update state transition function} \label{eq:state_transition} \\ 
   & p(R_{k+1} | R[Z_{k}], S\left[Z_{k}\right], A, q_k) \cdot    & \text{update reward function} \label{eq:reward_function} \\ 
   & p(\Omega_{k+1} | S\left[Z_{k}\right], q_k).   & \text{update partial observation function} \label{eq:rendering}
\end{align}

%\sun{explain more using the example in figure 2}
% Intuitively, \( Z_{k+1} \) is the subset of indices from the set \( \{1, 2, \ldots, n_k\} \) that refer to the relevant factors and states at step \( k+1 \). 
% 
% This allows us to, instead of having to generate everything from scratch, simply build on top of the existing POMDP.
%We make the observation that separating a prompt for Eq.~\ref{eq:reward_function} is unnecessary as it can be treated as part of the state transition function that updates a . 
Note that $Z_k$ can only consist of state variable indices in the state space $S_{k+1}$. %\sun{make it clear that z < i for all z \in Z_i$}
In practice, we achieve this by encouraging the LLM to select a minimal set of relevant states $Z_k$ in the prompt.

We find that the term~\ref{eq:state_transition} is most prone to error, likely because the most complicated functions of a simulation are state transitions. Motivated by this observation, we propose to adopt the \textit{model-view-controller} design pattern for structuring these prompts. Instead of prompting LLMs to update the state transition function first and then update the reward function, we prompt the LLMs to update the action-dependent part of the state transition function (i.e. the \textit{Controller} component) and then the action-independent part (i.e., \textit{Model}). We treat the reward model as part of the state transition function that updates a \textit{score} state variable. That is, $T(s^\prime | s, a) = T^{(s)}(s^\prime | s) T^{(a)}(s | s, a)$ where $T^{(a)}(s | s,a)$ denotes the part of the state transition function that handles how actions affect the states and $T^{(s)}(s^\prime | s)$ denotes the part of the state transition function that how states are updated every step. This gives us our final algorithm as illustrated in Algorithm 1.

In Algorithm 1, colors indicate the corresponding components of the model-view-controller pattern. Red highlights the \textit{controller}, corresponding to parts of the state transition dependent on user-input actions.%through action interactions.
Green shows the \textit{model}, corresponding to parts of the state transition function that are not dependent on user-input actions. 
Blue shows the \textit{view} component, updating the observation function that acts as the ``renderer'' of the state space.

\begin{table}[!t]
\caption{Percentage of system tests passed by different methods of generating 2D RL games.}
\small
\centering
\resizebox{\textwidth}{!}{%
\begin{tabular}{lcccccccc}
\toprule
\textit{\% of system tests passed.}& \multicolumn{1}{c}{\textbf{Flappy Bird}}   & \multicolumn{1}{c}{\textbf{Catcher}}  & \multicolumn{1}{c}{\textbf{Snake}}& \multicolumn{1}{c}{\textbf{Pixelcopter}}    & \multicolumn{1}{c}{\textbf{Pong}} & \multicolumn{1}{c}{\textbf{Puckworld}} & \multicolumn{1}{c}{\textbf{Waterworld}} & \multicolumn{1}{c}{\textbf{Monster Kong}} \\ 
%\cmidrule{2-15}
%\multicolumn{1}{l}{\textbf{Method}}  & \multicolumn{1}{c}{\# Token}  & \multicolumn{1}{c}{\# Token}  & \multicolumn{1}{c}{\# Token}  & \multicolumn{1}{c}{\# Token} & \multicolumn{1}{c}{Success} & \multicolumn{1}{c}{\# Token}  & \multicolumn{1}{c}{\# Token}  & \multicolumn{1}{c}{\# Token}\\ 
\midrule
Mistral-7B-Instruct & 0.00 & 0.00 & 0.00 & 0.00 & 0.00 & 0.00 & 0.00 & 0.00 \\
Llama-3                 & 0.15 & 0.33 & 0.19 & 0.14 & 0.01 & 0.43 & 0.25 & 0.29 \\
Llama-3 w/ self debug        & 0.15 & 0.41 & 0.28 & 0.19 & 0.03 & 0.44 & 0.22 & 0.31\\
Llama-3 CoT w/ self debug & 0.20 & 0.39 & 0.25 & 0.21 & 0.16 & 0.50 & 0.42 & 0.35\\
%Gemini-1.5 & \\
%Gemini-1.5 w/ debug &  \\
%Gemini-1.5 w/ debug w/ CoT & \\
GPT-3.5 & 0.19 & 0.39 & 0.37 & 0.38 & 0.22 & 0.33 & 0.34 & 0.19 \\
GPT-4 & 0.35 & 0.35 & 0.42 & 0.44 & 0.25 & 0.34 & 0.46 & 0.21\\
GPT-4 w/ self debug & 0.33 & 0.53 & 0.43 & 0.51 & \textbf{0.75} & 0.41 & 0.45 & 0.31\\
GPT-4 w/ AgentCoder & 0.18 & 0.45 & 0.27 & 0.43 & 0.43 & 0.33 & 0.20 & 0.23 \\
GPT-4 CoT w/ self debug & 0.30 &  0.51  & 0.39  & 0.53  & 0.64  & 0.47  & 0.50  & 0.34\\
\midrule
Llama-3 w/ \method{} (ours) & 0.55 & 0.54 & \textbf{0.50} & 0.41 & 0.38 & 0.58 &0.27 & 0.35 \\
GPT-4 w/ \method{} (ours) & \textbf{0.78}  & \textbf{0.66} & 0.44 & \textbf{0.78} &  0.61 & \textbf{0.81} &  \textbf{0.62} & \textbf{0.44} \\
%Gemini-1.5 w/ \method{} (ours) & \\
\bottomrule
\end{tabular}
}
\label{tab:code_gen_results}
%\vspace{-0.5cm}
\end{table}

\begin{figure}%{!br}{0.64\textwidth} % 'r' for right, '0.5\textwidth' for width of the figure
    \centering
    %\vspace{-4mm}
  \includegraphics[width=.8\linewidth]{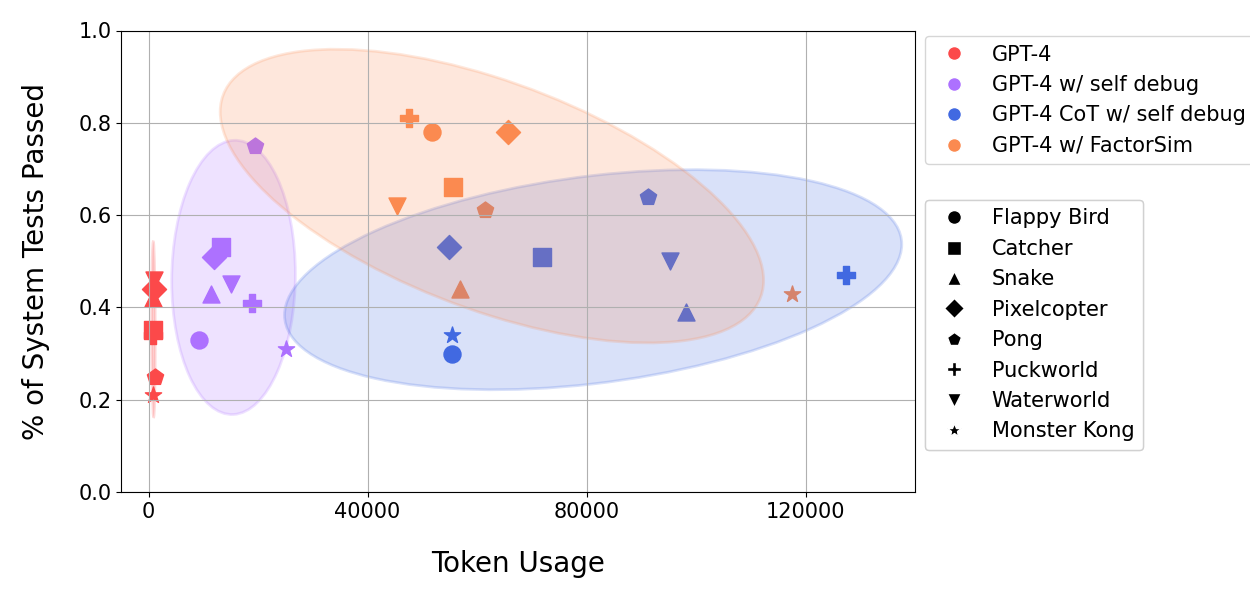}
   \caption{Performance and token usage analysis of GPT-4-based methods. Ellipses correspond to 90\% confidence intervals for each algorithm, aggregated over all RL games.}
  %\vspace{-2em}
    \label{fig:token}
\end{figure}

\section{Experiments}
In this paper, we consider two types of simulations: 2D Reinforcement Learning (RL) games and robotics tasks in a physics engine. We also introduce a new benchmark to evaluate generative simulation methods. Our experiments are designed to test three hypotheses. First, \method{} generates simulations with \textit{better prompt alignment}, which we evaluate through system tests and human evaluations. Second, \method{} enables \textit{better zero-shot transfer} by training RL agents in the simulated generated environments. Third, \method{}'s strengths in \textit{generating robotic tasks}.

\subsection{RL Game Generation}
%\paragraph{Experiment Setup}
%\paragraph{Talk about high-level}
% To answer \textbf{Q1} and \textbf{Q2}, we propose a new \textit{generative simulation} benchmark of RL games and evaluate our method against various baselines. We also conducted a human study.
% \paragraph{Benchmark Setup} 
To answer our first two hypotheses, we propose a new benchmark that includes all 2D games from the PyGame Learning Environment~\footnote{We exclude the sole 3D game Raycast Maze and leave 3D game generation to future work.}~\cite{tasfi2016PLE}: Flappy Bird, Catcher, Puckworld, Pixelcopter, Pong, Snake, Waterworld, and Monster Kong. 
% To create a \textit{generative simulation} benchmark, we need to have pairs of prompts and system tests
%~\footnote{The 3D game Raycast Maze is excluded due to its unique complexity not applicable to 2D games}.
% For each game, we first generate language descriptions by feeding the game's documentation to an LLM. 
% Each RL games' input prompt is the game's documentation found online; as most of the game documentations are not fully specified, we manually add details to documentations. This way, our method and baselines will not have to hallucinate any gaps in the specified game details, and will give all methods a fair evaluation. 
For each RL game, the input prompt consists of the game's online documentation. Since most game documentation is incomplete, we manually supplement them with additional details (see Appendix). This ensures that our method and the baselines do not hallucinate any missing game information, allowing for a fair evaluation across all methods.

% Due to the lack of detailed documentation and LLM hallucinations, we refined the prompts such that they reflect the original implementations. 
% Adopting common practices in game development, we use system tests that programmatically assess whether the generated simulations adhere to the logic specified. 
% These tests simulate actions such as key presses and mouse clicks and check whether the game states are changed correctly.
Following common practices in game development, we design system tests to verify that the generated simulations follow the specified logic programmatically. These tests simulate actions like key presses and mouse clicks and check if the game states are updated correctly. Refer to the Appendix for more details.

%Note that we opt for system tests instead of tests that can be run for individual game objects because that would effectively require the simulation to be of a specific structure. 
%We derive the system tests from the prompts for each game.

\paragraph{Baselines}
For baselines, we compare to three methods using a closed-source (GPT-4~\cite{achiam2023gpt}) and an open-source LLM (Llama-3~\cite{llama3modelcard}).
The first approach prompts the LLM with all contexts at once, which we denote as the \textit{vanilla} method.
The second approach uses \textit{self-debugging} \cite{chen2024teaching}, where the model retries generating the code when provided error messages from running the code (up to 10 times in our experiments).
A third approach combines \textit{Chain-of-Thought~\cite{wei2022chain} (CoT)} reasoning with self-debugging, where the LLM generates code incrementally, processing one instruction at a time. Additionally, we incorporate AgentCoder~\cite{huang2023agentcoder} as a baseline.
CoT with self-debugging is an ablation study of our method that acts without the factored POMDP representation.

\begin{figure}[!t]
  \centering
  \includegraphics[width=\linewidth]{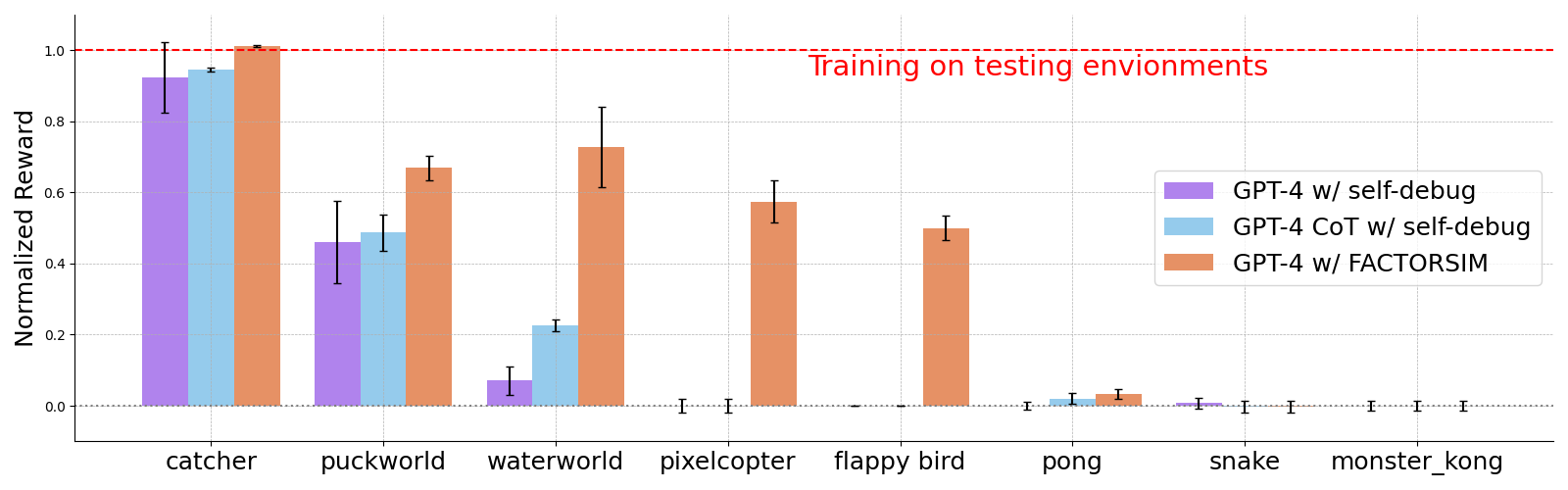}
   \caption{Zero-shot transfer results on previously unseen environments (i.e., environments in the original RL benchmark~\cite{tasfi2016PLE}).}
   \label{fig:zero_shot_transfer}
\end{figure}

\paragraph{Code Generation Evaluation}
Table~\ref{tab:code_gen_results} shows the results for the baselines and our method. \method outperforms all baselines in 7 out of 8 games. Additionally, we compare performance and LLM token usage across various methods using GPT-4 (Figure~\ref{fig:token}). While the vanilla baseline uses the fewest tokens, it only achieves moderate accuracy. Additionally, combining Chain-of-Thought (CoT) reasoning with self-debugging results in the highest token usage but only marginally improves accuracy over iterative self-debugging. 
%
% As shown in Figure~\ref{fig:token}, the one-shot baseline uses the fewest tokens and achieves moderate accuracy. 
% Adding a self-debugging mechanism increases both token usage and accuracy.
%
% Combining Chain-of-Thought (CoT) reasoning with self-debugging results in the highest token usage but only marginally improves accuracy over iterative self-debugging. 
%
\method achieves the highest accuracy with modest token usage, indicating that the decomposition of tasks reduces the need for extensive iterative debugging.

Empirically, we find that certain prompts, when tested on baselines without the decomposition mechanism in \method, are prone to syntax or runtime errors that the LLMs cannot self-debug. This is particularly evident with Llama-3 (vanilla) and Llama-3 self-debug, which perform poorly as they
% because they ignore the prompt specifications and repeatedly 
generate highly similar incorrect implementations, ignoring the logic specified in the prompts even when the temperature is set to 1. We hypothesize that this behavior is due to the model having a strong prior for how certain games, like Pong and Flappy Bird, should be implemented, ignoring the prompt specifications. This ``mode collapse'' issue of LLMs could be caused by over-fitting in the instruction tuning stage~\cite{hamilton2024detecting}.

While AgentCoder iteratively refines code, it performs poorly because it relies on a test designer agent and a test executor agent to write quality test cases. However, due to the complexity of the tasks in our benchmark, the test designer agent tends to write incorrect or infeasible tests, leading to negative feedback. This points to FactorSim being an improvement over the standard "role-based" Chain of Thought decompositions, and that it is non-trivial to generate simulations from complex textual specifications.
%\sun{even though the decomposition is at a finer-grained level, because the subtasks are easier and we need not as much iterative debugging, so we actually end up using less tokens compared to the Cot baseline}. \sun{talk about why vanilla llama baselines do worse than vanilla gpt in some games but these results are not symmetric in factorsim with llama and factorsim with gpt. Why does factorsim help llama by a wider margin in the snake game? - sometimes LLMs collapse/stubborn with certain implementation and will just not following the prompt (like in the case of pong) and just keep "reverting" back to particular implementation (maybe mention memorization)-- facotrization also helps with these cases}

\paragraph{Zero-shot Transfer Results} Additionally, we test \method by training a PPO~\cite{schulman2017proximal} agent on 10 generated environments for 10 million steps and zero-shot test it on the ``ground-truth'' environment implemented in the original RL benchmark (Figure~\ref{fig:zero_shot_transfer}).
%This ``ground-truth'' is available because the prompts are created based on the implementation in the original RL benchmark. 
% The results are shown in Figure~\ref{fig:zero_shot_transfer}. 
The rewards are linearly scaled such that 0 corresponds to the performance of a random policy and 1 corresponds to the performance of a PPO agent trained for 10 million steps on the "ground-truth" environment. \method achieves notably better zero-shot transfer results as a result of generating code that adheres more closely to the prompt specification. We also observe that the errors \method made tend to be more spread out across different components of the simulation. In contrast, many baselines suffer from failure modes concentrated in a specific aspect of the generation (e.g., incorrectly implementing the collision logic) that significantly hampers the ability of a set of generations to facilitate zero-shot transfer.\looseness=-1
%of the generated simulation.
%\sun{say more about why facotirzed method can perform better because of better prompt alignment}

%Figure 2 presents a comparison of human evaluation results for generated simulations from FactorSim and CoT Baseline across eight games. The simulations are categorized as fully playable, playable with issues, unplayable but rendered properly, and unplayable (crashes or fails to launch).
%\sun{one of the reasons that we ran human study is also beause our systems tests don’t test for the rendering functions. In our RL expeirments, the rendering function is fixed to be an identity function (we train on state variables not in pixel space}f
%FactorSim shows better performance in generating functional game simulations.
%It has 28.7\% of simulations fully playable and 38.9\% playable with issues, indicating a higher success rate in creating operational game environments. However, 21.6\% of FactorSim's simulations are unplayable but rendered properly, and 10.8\% crash or fail to launch, pointing to areas needing improvement. In contrast, CoT Baseline has only 21.2\% fully playable simulations and 24.7\% playable with issues, suggesting lower effectiveness. The major issue with CoT Baseline is its high rate of crashes or launch failures at 42.9\%, indicating significant stability problems. Only 11.2\% are unplayable but rendered properly.
%Overall, FactorSim outperforms CoT Baseline and demonstrates better performance in generating functional game simulations.

%Additionally, we show that \method generates simulations that are more ``playable'' by humans.
\begin{figure}[!t] %{!tr}{0.5\textwidth} % 'r' for right, '0.5\textwidth' for width of the figure
    \centering
    %\vspace{-4mm}
    \includegraphics[width=0.6\textwidth]{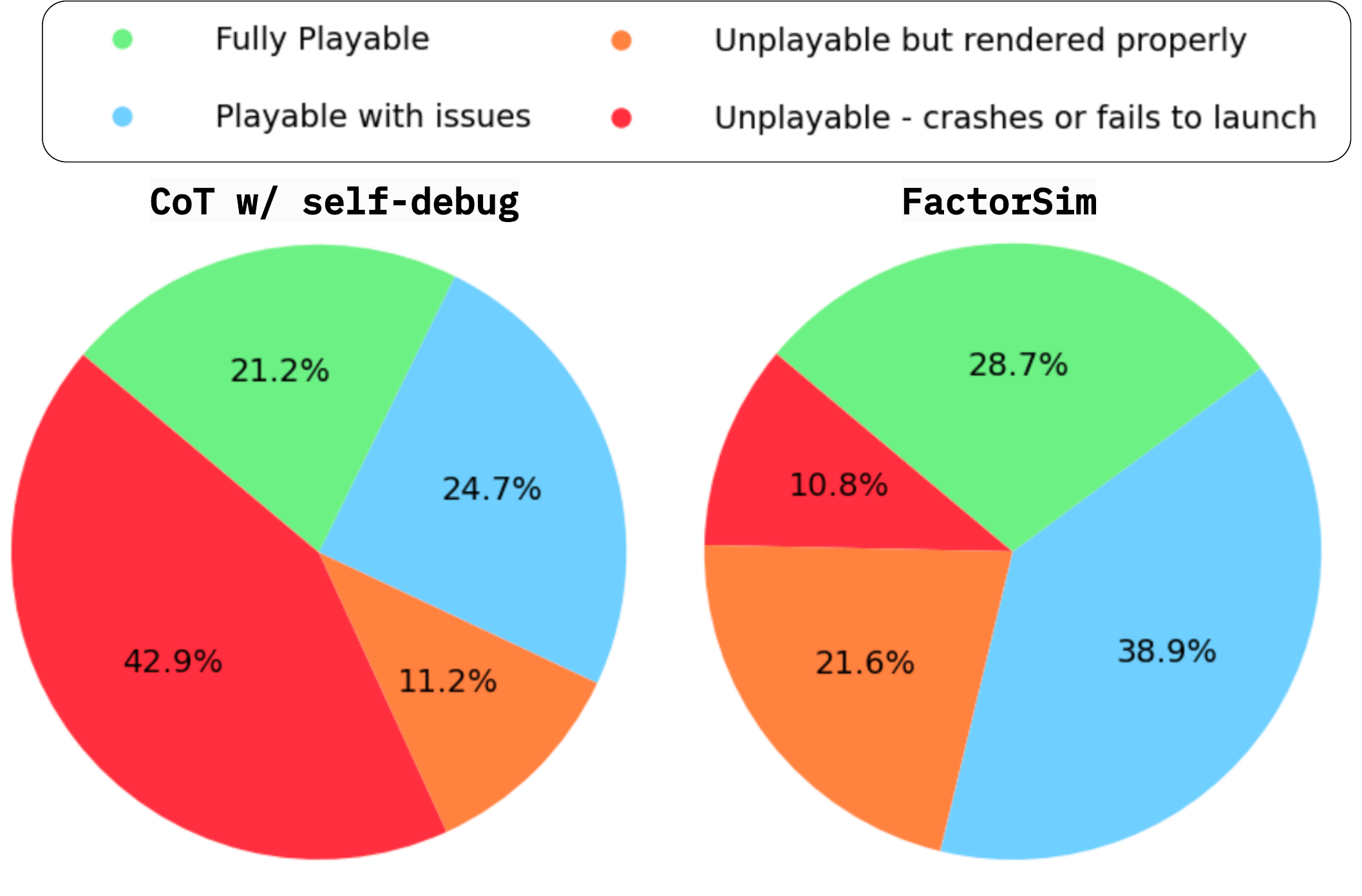}
    \caption{Human evaluation results on the generated simulations of \method and the strongest baseline (i.e., GPT-4 CoT w/ self-debug), aggregated over all 8 RL games.}
    %\vspace{-5mm}
    \label{fig:human_study_result}
\end{figure}
\paragraph{Human Study Evaluation} 
Automated systems tests cannot holistically capture some aspects of game playability such as rendering a usable user interface.
To address this limitation we conducted a human study where users were asked to play the generated games and evaluate their playability.
Over 320 human evaluations (40 per game) we find \method generates more functional and playable games, compared to the strongest baseline GPT-4 CoT with iterative self-debugging (Figure~\ref{fig:human_study_result}).
% To more comprehensively evaluate the generated RL games, we conducted a human study to assess aspects of the generation that system tests cannot holistically evaluate, such as the pixel rendering function of the simulations. We collected a total of 320+ human game-play evaluations (40+ per game).
% \method generates more functional and playable games, compared to the strongest baseline GPT-4 CoT with iterative self-debugging  (Figure~\ref{fig:human_study_result}). 
More details can be found in the Appendix.

\subsection{Robotics Task Generation}
\begin{figure}[!t]
  \centering
  \begin{subfigure}[b]{0.48\linewidth}
    \centering
    \includegraphics[width=\linewidth]{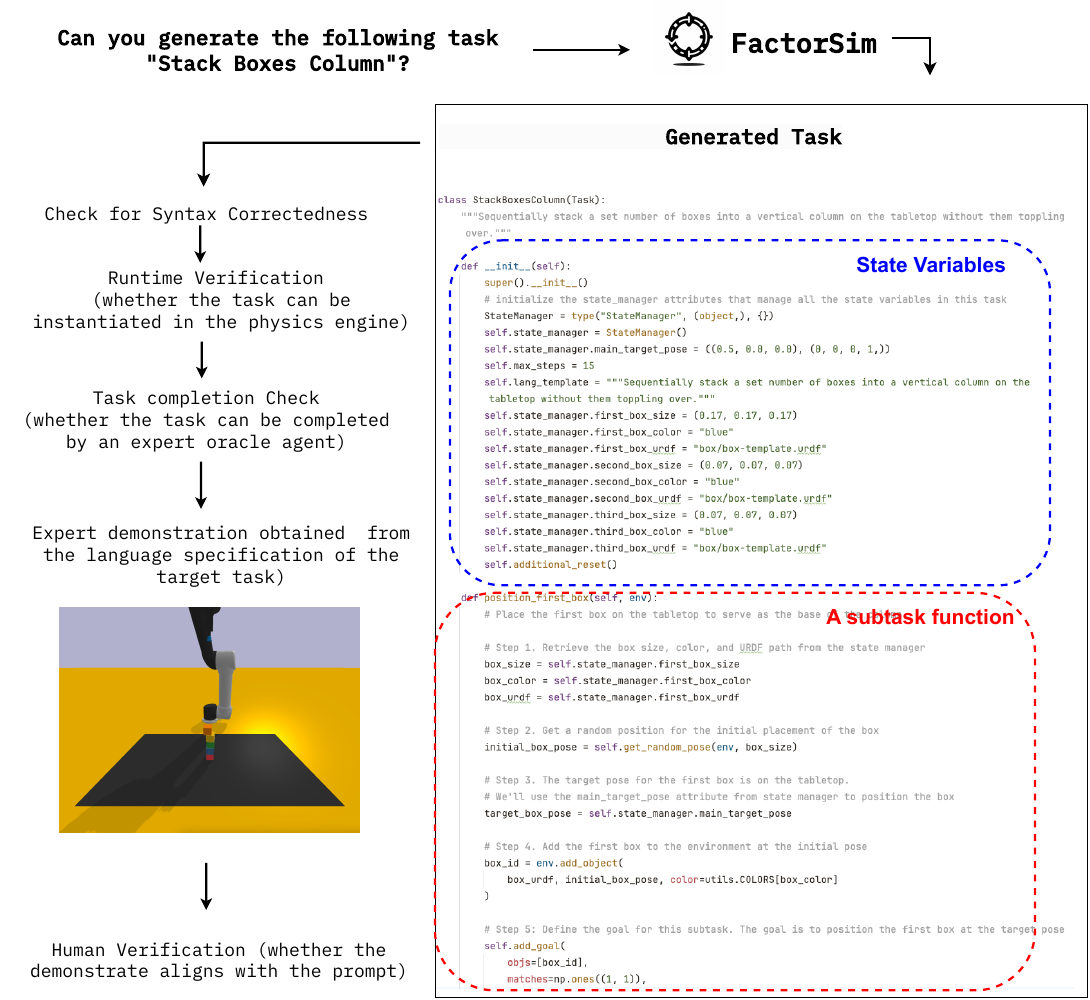}
    %\caption{Overview of our robotics task generation experimental setting.}
    %\label{fig:subfig1}
  \end{subfigure}
  \hfill
  \begin{subfigure}[b]{0.48\linewidth}
    \centering
    \includegraphics[width=0.95\linewidth]{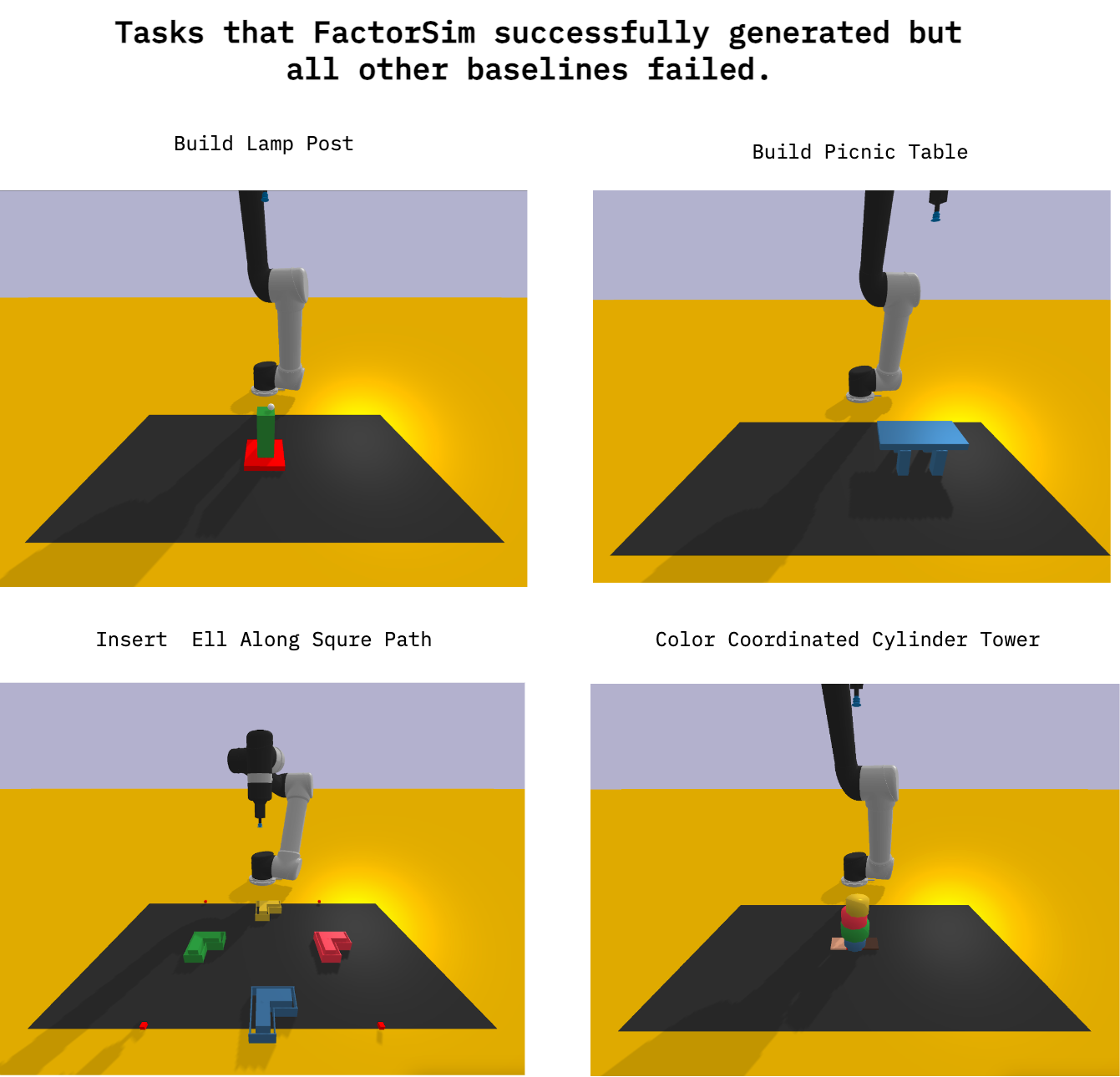}
    %\caption{}
    %\label{fig:subfig2}
  \end{subfigure}
  \caption{\textbf{Left}: an overview of our robotics task generation experimental setting. \textbf{Right}: Tasks successfully generated using FactorSim, which all other baselines fail on.}%Clarification on how we evaluate FactorSim on generating robotics tasks and demonstrations for FactorSim's applicability to Embodied AI.}
  \label{fig:robotics_overview}
\end{figure}

We evaluate on GenSim's~\cite{wang2023gensim} 50-task benchmark of robotics tasks in the CLIPort framework~\cite{shridhar2021cliport}. Refer to Figure~\ref{fig:robotics_overview} for an overview of our experimental setting.
% We use the 50-task benchmark GenSim~\cite{wang2023gensim} published alongside their code to benchmark \method against other methods for generating robotics in the CLIPort framework~\cite{shridhar2021cliport}. 
We compare \method{} with the best-performing methods in generating code that specifies tasks (object states and reward structure) that can be used to train robots. Analogous to the game generation experiment, we use \method to modularize the code generation process into subtasks and have LLMs generate each subtask using only a set of necessary states as context.  More details can be found in the Appendix.
% Thus the state transitional model $T$, and the observation model $\Omega$ are fixed. A task in this context is defined by an initial state distribution and a reward function that specifies the target state distribution.
%so in this case, what we are factorizing is the function that maps start states to end states (not state variables at the next step but the goal state variables) 

% \sun{TODO: why did we do and design this experiment?}
% \sun{Empirically, we found that task completion is imperfect because ...(a) it does not reflect whether the generated task matches the input prompt and also (b) the ``oracle'' is not capable of solving more complicated tasks even when the task specification is correct / because it's rule-based}

%We approached the problem by breaking down
%complex tasks into modular and composable pieces with the minimal context
%needed. Given a complex task, we decomposed the toplevel task into the shared
%state variables and independent subtasks. By reducing the complexity of the task
%into subtasks with a limited set of states, we decrease the possibility for
%errors on the later generated code produced by the foundational model. The
%decomposition attempts to exploit the world knowledge from LLMs and %reduces the
%need for referencing any existing task library. An example showing the breakdown of the tasks can be found in Figure ~\ref{fig:robotics_figure}.

%Prompting was done on a pre-trained gpt-4-1106-preview and
%meta-llama-3-70b-instruct model. A diagram of the workflow can be found in
%Figure~\ref{fig:robotics_workflow}.

%\paragraph{Baselines \& Metrics}
\paragraph{Baselines \& Metrics}
%To assess the effectiveness of the factorized representation,  that they released and benchmark our method against their best performing baselines.
We compare our method with the multiple GenSim baselines: vanilla (one-shot), Chain-of-Thought (topdown), and Chain-of-Thought (bottom-up). Adopting the same set of metrics, we evaluate all methods on a sequence of pass rates on ``syntax correctness'', ``runtime-verified'', and ``task completed''.
A generated task is considered ``completed'' if a coded oracle agent could collect 99\% of the total reward half of the time. 
% could achieve the goals specified in the generated task. Following GenSim's evaluation metric, an episode was considered "successful" when the agent achieved a total reward of 99%, and a task was considered "completed" when at least half of the total run episodes were successful.

We empirically found that the "task completion rate" is an imperfect metric for evaluating the success of a generated task. A task deemed "complete" by the oracle agent may fail to adhere to the prompt. For example, when asked to generate a task "build a wheel," a method might produce a task specification that involves rearranging blocks into a structure that does not resemble a wheel. To address this, we introduced a metric of the ``human pass rate''. This involved manually inspecting runtime-verified tasks to determine if the task specifications aligned with the prompt descriptions (see Appendix).

\begin{figure}[!t]
  \centering
  \includegraphics[width=\linewidth]{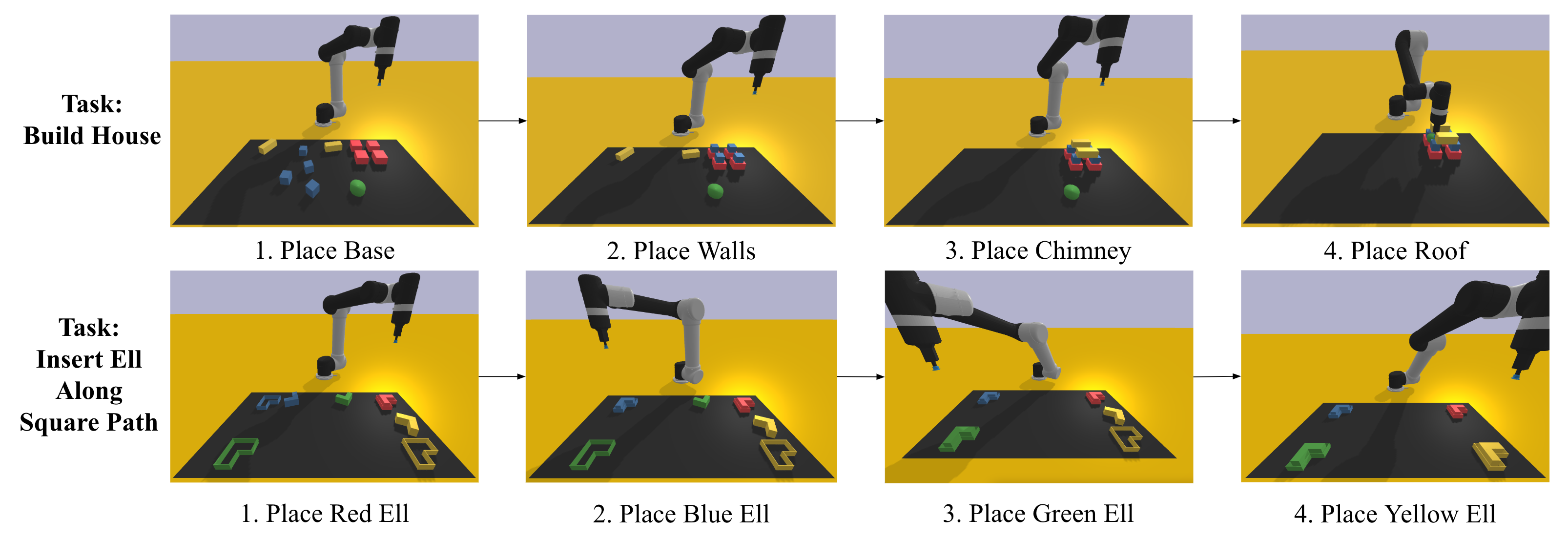}
   \caption{This figure illustrates two input task prompts and the corresponding sequence of subtasks decomposed by \method.}
   \label{fig:robotics_demo}
\end{figure}

\paragraph{Results \& Discussion}
\begin{figure*}[!t]
    \centering
    \vspace{-4mm}
    \includegraphics[width=\textwidth]{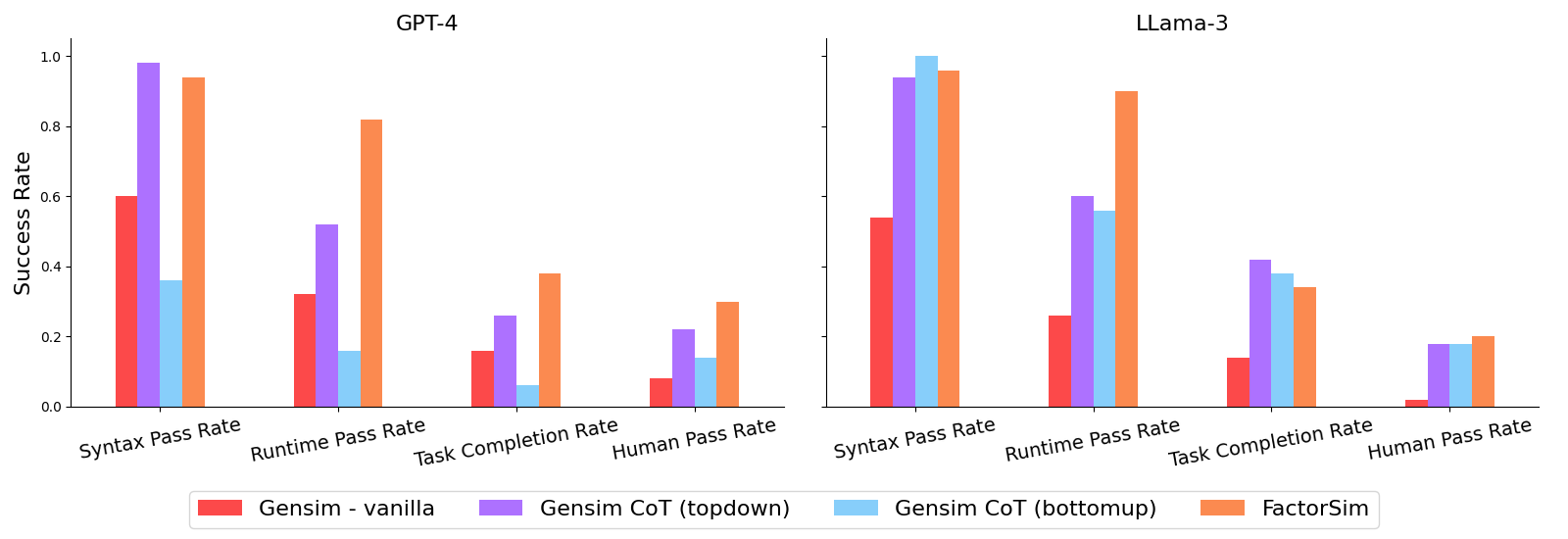}
        \vspace{-4mm}
    \caption{Performance of \method and GenSim~\cite{wang2023gensim} baselines in generating robotic tasks.}
    \vspace{-4mm}
        \label{fig:robotics_eval_llama}
\end{figure*}
\method outperforms baselines in generating tasks with a higher runtime pass rate and better human evaluation scores, indicating improved prompt alignment (Figure~\ref{fig:robotics_eval_llama}). Task completion rates are generally low for all methods due to the limitation of the oracle agent. For example, tasks like "Build Ball Pit" (fill a container with balls to create a ball pit) often fail because the balls roll out of the visible area of the oracle agent, not because the generated task is invalid. \method performs particularly well on tasks that specify spatial relationships (e.g., "on top of," "left of," "outside of") between objects, such as the "build House" example in Figure~\ref{fig:robotics_demo}. This improvement is likely due to the decomposition process, where for each step, instead of addressing a combination of multiple spatial relations all at once, \method attends to a smaller context, allowing each spatial relation to be addressed separately.

%\newpage

%\subsection{Scene Rearrangement/Generation}
%\sun{I have not decided whether to include this section or not}

%Factor graphs generalize constraint graphs
 
%Our goal is to approximate/model real-world layouts $p({X’_i} | { X_i }, q_\text{text})$ 

%$q_\text{text})$  is the human instruction 
%$X_i$ is the random variable that represents the 6D pose of the object $x_i$ 

%Data-driven approaches directly model p, but doing this for open-vocabulary is insanely hard

%Exploit the structure/modularity in real-world scenes!
%i.e. $p({X’_i} | { X_i }, q_\text{text}) = f_1({X_1, X_3} | q_\text{text}) * f_2({X_1, X_2} | q_\text{text}) * f_3(...)$

%How do I use a foundational model to decide how to factorize this probability distribution?

%\subsection{RL experiments}
%Can we utilize this LLM-inferred factorization technique to train better RL agents? 

%\vspace{-2mm}

\section{Conclusion \& Future Work}
%\vspace{-2mm}
We have proposed \method as an approach to generate full simulations as code that can train agents while adhering to detailed design requirements specified as a text prompt. We also introduce a benchmark suite of eight RL environments to evaluate generative simulation methods.% that includes both system tests to verify whether a coded simulation aligns with prompt specifications and tests verifying RL agents trained in the simulations generalize to unseen environments. 

Generating complex simulations in code is challenging, and we anticipate numerous opportunities to extend the simulation generation process. 
% Future work could focus on incorporating feedback from the agent training process while simultaneously adapting the entire generation process to enhance agent training. 
%Generating complex simulations in code is challenging and we anticipate many avenues of work to address of the limitations of \method.
% simulators
There is substantial room to address larger-scale, more complex games, and robotics environments that require code bases beyond what can be used effectively in the context window of existing LLMs.
We also see great potential to accelerate RL agent training by generating code that can be accelerated on GPU devices. % and to generate fully differentiable simulators.
Our robotic simulation results will benefit from further investigations to demonstrate transfer to real-world environments.
% agents
We have only addressed single-agent simulations, leaving the extension of our method to multi-agent settings to future work. In the future, we also plan to incorporate information from the agent training process to automatically modify the generated simulation environment for enhanced agent learning and generalization. 
Taken together, we believe the generation of full simulations as code will be an important step toward enhancing the capabilities of LLMs to support the development of generalized RL agent policies.

%\section{Broader Implication}
%In the appendix, we discuss more limitations of our method.

% limitations / future work (\zook{initial thoughts}):
% \begin{itemize}
%     \item not differentiable (vs neural world models)
%     \item not (currently) code that accelerates on GPU/TPU
%     \item small scale of games / environments
%     \item not automatically improving environments to train agents. aka environment design is not (yet) iterative. also no curriculum aspect.
%     \item want to show robotic sim2real
%     \item single agent only. want to extend method and factor graph formalization to handle multi-agent case
% \end{itemize}

\begin{ack}
This work was in part supported by the Stanford Institute for Human-Centered Artificial Intelligence (HAI), the Stanford Center for Integrated Facility Engineering (CIFE), NSF CCRI \#2120095, AFOSR YIP FA9550-23-1-0127, ONR N00014-23-1-2355, ONR YIP N00014-24-1-2117, and Google.
\end{ack}

%$\bibliography{example_paper}
%$\bibliographystyle{plain}
\newpage
\bibliographystyle{plain}
\bibliography{neurips_2024}

%%%%%%%%%%%%%%%%%%%%%%%%%%%%%%%%%%%%%%%%%%%%%%%%%%%%%%%%%%%%
%%%%%%%%%%%%%%%%%%%%%%%%%%%%%%%%%%%%%%%%%%%%%%%%%%%%%%%%%%%%
\clearpage
\appendix

%Optionally include supplemental material (complete proofs, additional experiments and plots) in appendix.
%All such materials \textbf{SHOULD be included in the main submission.}

%\section{Appendix / supplemental material}
%\section{Limitations \& Future Work}

\section{Societal Impact}
This work can be applied broadly to many types of simulations, including robotics, autonomous vehicles, and other autonomous systems.
Such systems have the potential for both positive and negative societal impact (e.g., harmful dual use).
As researchers, we must critically evaluate such applications and promote beneficial ones.
In this work we have focused on simulations with potential positive social impact, particularly in supporting the development of robots able to operate in human environments like households or manufacturing facilities.

The methods we present generate simulations that can be used to train agents to perform tasks.
One risk with generated simulations is for training agents in an unintended manner.
By generating simulations specified as code we mitigate this concern by making the behavior of the simulation explicit and inspectable by humans.
Further, our approach is better able to guide LLMs to generate code that matches input design specifications compared to baseline methods, reducing the risk of LLMs inadvertently producing undesirable functionality.
We believe this can help enhance the reliability of generated simulations while offering strong editing and control capabilities to humans.

The potential negative environmental impact of the compute for using our technique is small.
We have shown our technique consumes less tokens than comparable methods to yield equally good results.
Thus our method can be seen as a way to reduce computational needs when using LLMs for tasks like creating simulations.
Compared to systems that use a neural world model our approach benefits from the relatively lower computational costs of running simulations in code compared to running large neural models for simulation.

%\section{More Related Work}
%\sun{TODO}

\section{Additional details of our experiments}
\label{sec:experiment_details}
All experiments are done on a workstation with 8 Nvidia A40 GPUs and
1008G of RAM. For our code generation experiments, one generation (i.e., generation of one training environment) takes around 30 seconds to 5 minutes. For our Reinforcement Learning experiments, one trial of training (i.e. training on a set of environments for 10M steps in total) takes around 3-5 hours to complete.
In all of our experiments, GPT-4 refers to the OpenAI's ``gpt-4-1106-preview'' model, GPT-3.5 refers to OpenAI's ``gpt-3.5-turbo'' model, and Llama-3 refers to the open-sourced ``meta-llama-3-70b-instruct'' model that can be found on huggingface.

For the zero-shot transfer RL experiment, we supply all methods with the reference "controller" (i.e., the same key press/mouse click leads to the same thing). We do this because language descriptions of such can be very ambiguous (e.g., a description ``a key press leads the bird to flap its wings'' can imply a change in position, velocity, or acceleration). % Zero-shot generalization bef an action does not mean the same thing in the generated simulations as in the target simulations, generalization becomes difficult. 
In our experiments, we generate 10 environments and filter out those that cannot be properly executed with a random policy. 

% \section{Additional details for the RL agent training experiment}

% used PPO with Ray implementation
% Parameters
% 2048 SGD minibatch
% rollout fragment length 1000
% training batch size 10000
% LSTM:
% - fcnet hidden 4,4
% - ReLU for CNN, FC, post-FCnet
% - cell size 64
% - use previous action, not previous reward
% 10M steps
% see code for further details
All the RL experiments are implemented in RLLib~\cite{liang2018rllib}~\footnote{https://docs.ray.io/en/latest/rllib/index.html}. The PPO agent is trained with a batch size of 10,000, and an SGD minibatch size of 2048. Our agent used a fully connected network with hidden layers of sizes (4, 4) and post-FCNet hidden layers of size 16, all employing ReLU activation functions. The policy network uses an LSTM with a cell size of 64 to incorporate previous actions but not previous rewards. Over the course of the 10 million training steps, 20 checkpoints were saved, with the best zero-shot performance on the testing environment reported.

\section{Additional details of \method}
We provide code in the supplementary material. Here we provide the prompts used in \method.
\begin{lstlisting}[label={list:supp:prompts:1},language={},caption=The first decompositional prompt used in \method.]
Given an unstructured game description, decompose the game's specification into a set of steps or modules. Each step or module should contain at most one input event handling, one state transitional logic, and one rendering logic.
If we model the game as a Markov Decision Process (MDP), the steps, or modules of the game, should share as little state variables as possible.

Please provide the response in the following format:
```json
{{
    "steps": [
        "Step 1: Describe the first module/step",
        "Step 2: Describe the second module/step",
        "Step 3: Describe the third module/step",
        ...
    ],
    "explanations": "To make sure the decomposition don't miss any important game mechanics, explain where each module/step fits in the game's logic."
}}
```

For example:
```json
{{
    "steps": [
        "Step 1: Introduce a balloon asset rendered as a blue rectangle. Implement gravity so that the balloon is by default always falling down. Allow users to move the balloon with arrow keys.",
        "Step 2: Implement a scoring system that rewards the human player for returning the ball back (i.e. making contact with the ball). Display the current score at the top of the screen. Display a 'Game Over!' message when one player reaches a score of 5 and provide an option for the player to restart the game after the 'Game Over' screen is displayed."
        ...
        "Step 5: Implement a 'Game Over' screen that is displayed when the player reaches a score of 5. Allow the player to restart the game from this screen."
    ],
    "explanations": "The balloon asset is the main character of the game. Its rendering and movement are handled in the first module. The scoring system is implemented in the second module to keep track of the player's progress. The 'Game Over' logic is implemented in the last module to provide a clear end to the game."
}}

The unstructured specification of the game is:
\{game_specification\}


Please provide the structured steps in the format of the JSON above. 
- Ensure that each step is a separable part of the game that can be implemented as independently as possible.
- You most likely don't need to decompose the game into more than 5 steps. However, the most important thing is to ensure that all the steps accurately describe the game's implementation.
- The most important thing is to make sure that the decomposition does not miss any logic step (e.g., the balloon should not be able to go off the screen).
- Note that the order of the steps is the order that these modules will be called in the game loop. Ensure that the game described can be implemented sequentially. For example, the reset position logic should be implemented after the collision detection logic.
  
\end{lstlisting}

\begin{lstlisting}[label={list:supp:prompts:2},language={},caption=The state context selection prompt used in \method.]
The game designer is building a single-player game in Pygame by modeling this game as a Markov Decision Process (MDP).
Your task is to identify and compile a list of relevant state variables to implement a specific feature requested by the game designer.
The game already has the following state space implementation:
```python
import pygame
import sys
import random


{state_manager_code}

    # new variables will be added here:
    # variable_description
    self.{{variable_name}} = {{variable_type}}({{value}})
```

Please provide the state variables in the following format within a JSON object:
```json
{{
    "relevant_state_variables": [
        {{
            "variable_name": "Name of the variable",
        }},
        ...
    ],
    "new_state_variables": [
        {{
            "variable_description": "Description of the variable",
            "variable_name": "Name of the variable",
            "variable_type": "Type of the variable, one of {{int, float, str, bool, tuple, list}}",
            "variable_value": "Value of the variable, e.g. 100, 0.5, 'balloon', tuple((255, 0, 255)), True, [10, 50], [{{'x': 100, 'y': 200}}]",
        }},
        ...
    ]
}}
```

```

The game designer's request is: {query}.

Here are the dos and don'ts for this request:
- The list "relevant_state_variables" should contain the names of the existing state variables that are relevant to the game designer's request.
- Please return a single list of state variables that contains both existing variables that you think are relevant and new state variables.
- A software engineer will later implement this request by implementing a function that takes these variables as input, so ensure all the variables needed to implement the request are included.
- It is okay to include variables that don't end up being used in the implementation because redundant state variables will be filtered out later.
- Please provide all rendering variables (e.g., size, color) if there are components to be rendered. Color should never be white since the background is white.
- Don't provide Sprite, Surface, or Rect variables. We will handle these variables later.
- Don't introduce variables using existing variables (e.g., self.bird_size = self.pipe_size/2), all state variables should be independent of each other.
- Always provide a default value even if a state variable should be chosen randomly. The randomness will be implemented later.
- "variable_value" should never to empty like []. Always provide a non-empty default value so the software engineer can infer how the variable can be accessed.
- Do not hallucinate external image files (e.g., .png, .jpg) or sound effects(e.g., mp3).
- Prioritize reusing the existing state variables as much as possible. For example, if we have "position_x" and "position_y" of a character, do not give me another variable "positions" in a list format.
- Additionally, you may add new state variables to the list "new_state_variables" if necessary. Please only create new state variables if necessary.
"""
\end{lstlisting}

\begin{lstlisting}[label={list:supp:prompts:3},language={},caption=The prompt for the \textbf{Controller} component (as defined in the Model-View-Controller) utilized in the \method.]
The game designer is building a single-player game in Pygame by modeling this game as a Markov Decision Process (MDP). Your task is to detect key/mouse input and update the state variables accordingly according to a feature requested by the game designer.
The game has the following implementation already:
```python
import pygame
import sys
import random


{state_manager_code}

# existing input event handling functions
{existing_implementation}
# the new logic function will be here
# if the function is already implemented, it will be replaced with the new implementation

def main():
    state_manager = StateManager()
    running = True
    while running:
        event = pygame.event.poll()
        if event.type == pygame.QUIT:
            running = False
        # {{function_description}}
        {{function_name}}(state_manager, event)
    pygame.quit()

if __name__ == "__main__":
    pygame.init()
    main()
```

To ensure the implementation is correct, please also implmeent an unit test for the function. Please implement the following request from the game designer and return your answer in the following format:
```json
{{
    "function_name": "{function_name}",
    "function_description": "{function_description}",
    "function_implementation": "the pygame implementation of the function, including the first line of the function definition",
    "unit_test": "the unit test code for the function"
}}
```

Here are the dos and don'ts for this request:
- Note that the implementation of the function shuold only have two arguments (i.e. state_manager and event).
- The function implementation should involve checking user input with event (i.e. event.type and event.key).
- Minimize the number of functions added while meeting the game designer's requirements. However, make sure to always give the full implementation of the function.
- Include only the essential details requested by the game designer. Do not add things that are not requested.
- Please use the state variables defined in the state manager. Do not introduce new state variables.
- Only KEYDOWN events will be detected. Do not rely on KEYUP events.
- Check for index out of bounds errors with any lists or arrays being used.
- Check for divide-by-zero errors.
- Do not leave any code incomplete. Do not leave placeholder values. Do not provide demonstration code implementation. Be sure all code is fully implemented.
\end{lstlisting}

\begin{lstlisting}[label={list:supp:prompts:4},language={},caption=
The prompt for the \textbf{Model} component (as defined in the Model-View-Controller) utilized in the \method.]
The game designer is building a single-player game in Pygame by modeling this game as a Markov Decision Process (MDP). Your task is to define and code new state transition functions according to the feature requested by the game designer.
The game has the following implementation already:
```python
import pygame
import sys
import random


{state_manager_code}

# existing state transition functions
{existing_implementation}
# the new function will be here
# if the function is already implemented, it will be replaced with the new implementation


def main():
    state_manager = StateManager()
    running = True
    while running:
        event = pygame.event.poll()
        if event.type == pygame.QUIT:
            running = False
        # {{function_description}}
        {{function_name}}(state_manager)
    pygame.quit()

if __name__ == "__main__":
    pygame.init()
    main()
```

To ensure the implementation is correct, please also implmeent an unit test for the function. Please implement the following request from the game designer and return your answer in the following format:
```json
{{
    "function_name": "{function_name}",
    "function_description": "{function_description}",
    "function_implementation": "the pygame implementation of the function, including the first line of the function definition",
    "unit_test": "the unit test code for the function"
}}
```

Here are the dos and don'ts for this request:
- Only implement things that pertain to updating the state variables. Other aspects of the game like input event handling and UI components will be handled separately.
- Include only the essential details requested by the game designer. Do not add things that are not requested.
- These state transition functions will be called in every iteration of the main game loop. If you want to add a conditional logic to the function, please implement it in the function itself.
- Note that this new function will be added to the end of the list of state transition functions.
- Please use the state variables defined in the state manager. Do not introduce new state variables.
- Check for index out of bounds errors with any lists or arrays being used.
- Check for divide-by-zero errors.
- Do not leave any code incomplete. Do not leave placeholder values. Do not provide demonstration code implementation. Be sure all code is fully implemented.
\end{lstlisting}

\begin{lstlisting}[label={list:supp:prompts:5},language={},caption=The prompt for the \textbf{View} component (as defined in the Model-View-Controller) utilized in the \method.]
The game designer is building a single-player game in Pygame by modeling this game as a Markov Decision Process (MDP). Your task is to add rendering functions that decide how state variables are rendered as UI components on the screen, according to the feature requested by the game designer.
The game has the following implementation already:
```python
import pygame
import sys
import random


{state_manager_code}

# existing rendering functions
{render_code}
# the new function will be here
# if the function is already implemented, it will be replaced with the new implementation

def main():
    state_manager = StateManager()
    clock = pygame.time.Clock()
    running = True
    while running:
        action = pygame.event.poll()
        if action.type == pygame.QUIT:
            running = False

        # all the code for state transitional logics
        # omitted for brevity

        # Fill the screen with white
        state_manager.screen.fill((255, 255, 255))
        # all the code for rendering states as UI components
        # {{function_description}}
        {{function_name}}(state_manager)
        pygame.display.flip()
        state_manager.clock.tick(state_manager.fps)

    pygame.quit()

if __name__ == "__main__":
    pygame.init()
    pygame.display.set_caption("")
    main()
```

To ensure the implementation is correct, please also implmeent an unit test for the function. Please implement the following request from the game designer and return your answer in the following format:
```json
{{
    "function_name": "{function_name}",
    "function_description": "{function_description}",
    "function_implementation": "the pygame implementation of the function, including the first line of the function definition",
    "unit_test": "the unit test code for the function"
}}
```

Here are the dos and don'ts for this request:
- Only implement things that pertain to how state variables are rendered as UI components on the screen. Other aspects like input event handling and state transition will be handled separately.
- Please make sure that all of the state variables remain unchanged in the rendering functions.
- Include only the essential details requested by the game designer. Do not add things that are not requested.
- These rendering functions will be called in every iteration of the main game loop. If you want to add a conditional logic to the function, please implement it in the function itself.
- Note that the background color of the screen is white so white UI components will not be visible. Do not fill the screen with white again in the rendering functions.
- Note that the new function will be added to the end of the list of rendering functions.
- Please use the state variables defined in the state manager. Do not introduce new state variables.
- Check for index out of bounds errors with any lists or arrays being used.
- Check for divide-by-zero errors.
- Do not call pygame.display.set_mode in UI functions. Only call it once outside of any UI function that is called multiple times.
- Do not leave any code incomplete. Do not leave placeholder values. Do not provide demonstration code implementation. Be sure all code is fully implemented.
\end{lstlisting}

\section{Additional details and results for the robotics task generation experiment}
In this section, we provide the prompts we used for \method in the robotics task generation experiment. The prompts for the baselines can be found in the GenSim paper\footnote{https://github.com/liruiw/GenSim}~\cite{wang2023gensim}.% The workflow of our generation can be shown in Figure~\ref{fig:robotics_workflow}. 

To conduct human evaluation, we begin by observing the oracle agent attempting to solve the task. If the oracle agent successfully completes the task, we then assess whether the resulting goal states align with the input task prompt. If the oracle agent fails to solve the task, we investigate the reason for the failure. Often, the cause is apparent, such as the target container being too small or not having the right color of objects for the task. These are marked as failures. For cases where it is clear that the limitation lies in the oracle agent's ability, or when the reason for failure is not immediately apparent, we manually inspect the code for the task specification and base our decision on both the code and our observation of the oracle agent's attempt at solving the task.% If uncertainty persists, we inspect the code to verify the accuracy of the task specification and ensure it aligns with the prompt descriptions.

\begin{table}[]
\resizebox{0.95\textwidth}{!}{
\begin{tabular}{|c|cccc||cccc||cccc||cccc|}
\hline
                                       & \multicolumn{4}{c|}{\textbf{FactorSim}}                                                                                                 & \multicolumn{4}{c|}{GenSim - vanilla}                                                                                         & \multicolumn{4}{c|}{GenSim Chain of Thought (topdown)}                                                                         & \multicolumn{4}{c|}{GenSim Chain of Thought (bottomup)}                                                                        \\ \cline{2-17}
     \multirow{-2}{*}{\textbf{Target Tasks}} & \multicolumn{1}{c|}{Syntax}                & \multicolumn{1}{c|}{Runtime}               & Task & Human        & \multicolumn{1}{c|}{Syntax}                & \multicolumn{1}{c|}{Runtime}               & Task & Human       & \multicolumn{1}{c|}{Syntax}                & \multicolumn{1}{c|}{Runtime}               & Task & Human        & \multicolumn{1}{c|}{Syntax}                & \multicolumn{1}{c|}{Runtime}               & Task & Human        \\ \hline
%\multicolumn{1}{|c|}{BuildDogHouse}          & \multicolumn{1}{c|}{{\color[HTML]{11734B} yes}} & \multicolumn{1}{c|}{{\color[HTML]{B10202} no}}  & {\color[HTML]{B10202} no}  & \multicolumn{1}{c|}{{\color[HTML]{B10202} no}}  & \multicolumn{1}{c|}{{\color[HTML]{B10202} no}} & \multicolumn{1}{c|}{{\color[HTML]{B10202} no}}  & {\color[HTML]{B10202} no} & \multicolumn{1}{c|}{{\color[HTML]{B10202} no}} &\multicolumn{1}{c|}{{\color[HTML]{11734B} yes}} & \multicolumn{1}{c|}{{\color[HTML]{11734B} yes}} & {\color[HTML]{B10202} no}  & \multicolumn{1}{c|}{{\color[HTML]{B10202} no}} & \multicolumn{1}{c|}{{\color[HTML]{11734B} yes}} & \multicolumn{1}{c|}{{\color[HTML]{11734B} yes}} & {\color[HTML]{B10202} no}  \multicolumn{1}{c|}{{\color[HTML]{B10202} no}} &\\
BuildDogHouse                       & \multicolumn{1}{c|}{{\color[HTML]{11734B} yes}} & \multicolumn{1}{c|}{{\color[HTML]{B10202} no}}  & \multicolumn{1}{c|}{{\color[HTML]{B10202} no}}  & {\color[HTML]{B10202} no}  & \multicolumn{1}{c|}{{\color[HTML]{B10202} no}}  & \multicolumn{1}{c|}{{\color[HTML]{B10202} no}}  & \multicolumn{1}{c|}{{\color[HTML]{B10202} no}} & {\color[HTML]{B10202} no}  & \multicolumn{1}{c|}{{\color[HTML]{11734B} yes}} & \multicolumn{1}{c|}{{\color[HTML]{11734B} yes}} & \multicolumn{1}{c|}{{\color[HTML]{B10202} no}}  & {\color[HTML]{B10202} no}  & \multicolumn{1}{c|}{{\color[HTML]{11734B} yes}} & \multicolumn{1}{c|}{{\color[HTML]{11734B} yes}} & \multicolumn{1}{c|}{{\color[HTML]{B10202} no}}  & {\color[HTML]{B10202} no}  \\
BuildLampPost                       & \multicolumn{1}{c|}{{\color[HTML]{11734B} yes}} & \multicolumn{1}{c|}{{\color[HTML]{11734B} yes}} & \multicolumn{1}{c|}{{\color[HTML]{11734B} yes}} & {\color[HTML]{11734B} yes} & \multicolumn{1}{c|}{{\color[HTML]{11734B} yes}} & \multicolumn{1}{c|}{{\color[HTML]{B10202} no}}  & \multicolumn{1}{c|}{{\color[HTML]{B10202} no}} & {\color[HTML]{B10202} no}  & \multicolumn{1}{c|}{{\color[HTML]{11734B} yes}} & \multicolumn{1}{c|}{{\color[HTML]{B10202} no}}  & \multicolumn{1}{c|}{{\color[HTML]{B10202} no}}  & {\color[HTML]{B10202} no}  & \multicolumn{1}{c|}{{\color[HTML]{11734B} yes}} & \multicolumn{1}{c|}{{\color[HTML]{B10202} no}}  & \multicolumn{1}{c|}{{\color[HTML]{B10202} no}}  & {\color[HTML]{B10202} no}  \\
BuildNewsstand                      & \multicolumn{1}{c|}{{\color[HTML]{11734B} yes}} & \multicolumn{1}{c|}{{\color[HTML]{11734B} yes}} & \multicolumn{1}{c|}{{\color[HTML]{B10202} no}}  & {\color[HTML]{B10202} no}  & \multicolumn{1}{c|}{{\color[HTML]{11734B} yes}} & \multicolumn{1}{c|}{{\color[HTML]{B10202} no}}  & \multicolumn{1}{c|}{{\color[HTML]{B10202} no}} & {\color[HTML]{B10202} no}  & \multicolumn{1}{c|}{{\color[HTML]{11734B} yes}} & \multicolumn{1}{c|}{{\color[HTML]{B10202} no}}  & \multicolumn{1}{c|}{{\color[HTML]{B10202} no}}  & {\color[HTML]{B10202} no}  & \multicolumn{1}{c|}{{\color[HTML]{11734B} yes}} & \multicolumn{1}{c|}{{\color[HTML]{B10202} no}}  & \multicolumn{1}{c|}{{\color[HTML]{B10202} no}}  & {\color[HTML]{B10202} no}  \\
BuildBench                          & \multicolumn{1}{c|}{{\color[HTML]{11734B} yes}} & \multicolumn{1}{c|}{{\color[HTML]{B10202} no}}  & \multicolumn{1}{c|}{{\color[HTML]{B10202} no}}  & {\color[HTML]{B10202} no}  & \multicolumn{1}{c|}{{\color[HTML]{11734B} yes}} & \multicolumn{1}{c|}{{\color[HTML]{B10202} no}}  & \multicolumn{1}{c|}{{\color[HTML]{B10202} no}} & {\color[HTML]{B10202} no}  & \multicolumn{1}{c|}{{\color[HTML]{11734B} yes}} & \multicolumn{1}{c|}{{\color[HTML]{B10202} no}}  & \multicolumn{1}{c|}{{\color[HTML]{B10202} no}}  & {\color[HTML]{B10202} no}  & \multicolumn{1}{c|}{{\color[HTML]{11734B} yes}} & \multicolumn{1}{c|}{{\color[HTML]{B10202} no}}  & \multicolumn{1}{c|}{{\color[HTML]{B10202} no}}  & {\color[HTML]{B10202} no}  \\
BuildPicnicTable                    & \multicolumn{1}{c|}{{\color[HTML]{11734B} yes}} & \multicolumn{1}{c|}{{\color[HTML]{11734B} yes}} & \multicolumn{1}{c|}{{\color[HTML]{11734B} yes}} & {\color[HTML]{11734B} yes} & \multicolumn{1}{c|}{{\color[HTML]{B10202} no}}  & \multicolumn{1}{c|}{{\color[HTML]{B10202} no}}  & \multicolumn{1}{c|}{{\color[HTML]{B10202} no}} & {\color[HTML]{B10202} no}  & \multicolumn{1}{c|}{{\color[HTML]{11734B} yes}} & \multicolumn{1}{c|}{{\color[HTML]{11734B} yes}} & \multicolumn{1}{c|}{{\color[HTML]{B10202} no}}  & {\color[HTML]{B10202} no}  & \multicolumn{1}{c|}{{\color[HTML]{11734B} yes}} & \multicolumn{1}{c|}{{\color[HTML]{11734B} yes}} & \multicolumn{1}{c|}{{\color[HTML]{B10202} no}}  & {\color[HTML]{B10202} no}  \\
BuildBicycleRack                    & \multicolumn{1}{c|}{{\color[HTML]{11734B} yes}} & \multicolumn{1}{c|}{{\color[HTML]{B10202} no}}  & \multicolumn{1}{c|}{{\color[HTML]{B10202} no}}  & {\color[HTML]{B10202} no}  & \multicolumn{1}{c|}{{\color[HTML]{11734B} yes}} & \multicolumn{1}{c|}{{\color[HTML]{B10202} no}}  & \multicolumn{1}{c|}{{\color[HTML]{B10202} no}} & {\color[HTML]{B10202} no}  & \multicolumn{1}{c|}{{\color[HTML]{11734B} yes}} & \multicolumn{1}{c|}{{\color[HTML]{B10202} no}}  & \multicolumn{1}{c|}{{\color[HTML]{B10202} no}}  & {\color[HTML]{B10202} no}  & \multicolumn{1}{c|}{{\color[HTML]{11734B} yes}} & \multicolumn{1}{c|}{{\color[HTML]{B10202} no}}  & \multicolumn{1}{c|}{{\color[HTML]{B10202} no}}  & {\color[HTML]{B10202} no}  \\
BuildPicnicBasket                   & \multicolumn{1}{c|}{{\color[HTML]{11734B} yes}} & \multicolumn{1}{c|}{{\color[HTML]{B10202} no}}  & \multicolumn{1}{c|}{{\color[HTML]{B10202} no}}  & {\color[HTML]{B10202} no}  & \multicolumn{1}{c|}{{\color[HTML]{11734B} yes}} & \multicolumn{1}{c|}{{\color[HTML]{B10202} no}}  & \multicolumn{1}{c|}{{\color[HTML]{B10202} no}} & {\color[HTML]{B10202} no}  & \multicolumn{1}{c|}{{\color[HTML]{11734B} yes}} & \multicolumn{1}{c|}{{\color[HTML]{11734B} yes}} & \multicolumn{1}{c|}{{\color[HTML]{B10202} no}}  & {\color[HTML]{B10202} no}  & \multicolumn{1}{c|}{{\color[HTML]{11734B} yes}} & \multicolumn{1}{c|}{{\color[HTML]{11734B} yes}} & \multicolumn{1}{c|}{{\color[HTML]{B10202} no}}  & {\color[HTML]{B10202} no}  \\
BuildCylinderStructure              & \multicolumn{1}{c|}{{\color[HTML]{11734B} yes}} & \multicolumn{1}{c|}{{\color[HTML]{11734B} yes}} & \multicolumn{1}{c|}{{\color[HTML]{11734B} yes}} & {\color[HTML]{11734B} yes} & \multicolumn{1}{c|}{\color[HTML]{B10202} no}                         & \multicolumn{1}{c|}{{\color[HTML]{B10202} no}}  & \multicolumn{1}{c|}{{\color[HTML]{B10202} no}} & {\color[HTML]{B10202} no}  & \multicolumn{1}{c|}{{\color[HTML]{11734B} yes}} & \multicolumn{1}{c|}{{\color[HTML]{B10202} no}}  & \multicolumn{1}{c|}{{\color[HTML]{B10202} no}}  & {\color[HTML]{B10202} no}  & \multicolumn{1}{c|}{{\color[HTML]{11734B} yes}} & \multicolumn{1}{c|}{{\color[HTML]{B10202} no}}  & \multicolumn{1}{c|}{{\color[HTML]{B10202} no}}  & {\color[HTML]{B10202} no}  \\
BuildBridge                         & \multicolumn{1}{c|}{{\color[HTML]{11734B} yes}} & \multicolumn{1}{c|}{{\color[HTML]{11734B} yes}} & \multicolumn{1}{c|}{{\color[HTML]{11734B} yes}} & {\color[HTML]{B10202} no}  & \multicolumn{1}{c|}{{\color[HTML]{11734B} yes}} & \multicolumn{1}{c|}{{\color[HTML]{B10202} no}}  & \multicolumn{1}{c|}{{\color[HTML]{B10202} no}} & {\color[HTML]{B10202} no}  & \multicolumn{1}{c|}{{\color[HTML]{11734B} yes}} & \multicolumn{1}{c|}{{\color[HTML]{B10202} no}}  & \multicolumn{1}{c|}{{\color[HTML]{B10202} no}}  & {\color[HTML]{B10202} no}  & \multicolumn{1}{c|}{{\color[HTML]{11734B} yes}} & \multicolumn{1}{c|}{{\color[HTML]{B10202} no}}  & \multicolumn{1}{c|}{{\color[HTML]{B10202} no}}  & {\color[HTML]{B10202} no}  \\
BuildCar                            & \multicolumn{1}{c|}{{\color[HTML]{11734B} yes}} & \multicolumn{1}{c|}{{\color[HTML]{11734B} yes}} & \multicolumn{1}{c|}{{\color[HTML]{B10202} no}}  & {\color[HTML]{B10202} no}  & \multicolumn{1}{c|}{{\color[HTML]{11734B} yes}} & \multicolumn{1}{c|}{{\color[HTML]{11734B} yes}} & \multicolumn{1}{c|}{{\color[HTML]{B10202} no}} & {\color[HTML]{B10202} no}  & \multicolumn{1}{c|}{{\color[HTML]{11734B} yes}} & \multicolumn{1}{c|}{{\color[HTML]{11734B} yes}} & \multicolumn{1}{c|}{{\color[HTML]{11734B} yes}} & {\color[HTML]{B10202} no}  & \multicolumn{1}{c|}{{\color[HTML]{11734B} yes}} & \multicolumn{1}{c|}{{\color[HTML]{11734B} yes}} & \multicolumn{1}{c|}{{\color[HTML]{11734B} yes}} & {\color[HTML]{B10202} no}  \\
BuildTwoCircles                     & \multicolumn{1}{c|}{{\color[HTML]{11734B} yes}} & \multicolumn{1}{c|}{{\color[HTML]{11734B} yes}} & \multicolumn{1}{c|}{{\color[HTML]{B10202} no}}  & {\color[HTML]{B10202} no}  & \multicolumn{1}{c|}{{\color[HTML]{11734B} yes}} & \multicolumn{1}{c|}{{\color[HTML]{B10202} no}}  & \multicolumn{1}{c|}{{\color[HTML]{B10202} no}} & {\color[HTML]{B10202} no}  & \multicolumn{1}{c|}{{\color[HTML]{B10202} no}}  & \multicolumn{1}{c|}{{\color[HTML]{B10202} no}}  & \multicolumn{1}{c|}{{\color[HTML]{B10202} no}}  & {\color[HTML]{B10202} no}  & \multicolumn{1}{c|}{{\color[HTML]{B10202} no}}  & \multicolumn{1}{c|}{{\color[HTML]{B10202} no}}  & \multicolumn{1}{c|}{{\color[HTML]{B10202} no}}  & {\color[HTML]{B10202} no}  \\
BuildWheel                          & \multicolumn{1}{c|}{{\color[HTML]{B10202} no}}  & \multicolumn{1}{c|}{{\color[HTML]{B10202} no}}  & \multicolumn{1}{c|}{{\color[HTML]{B10202} no}}  & {\color[HTML]{B10202} no}  & \multicolumn{1}{c|}{{\color[HTML]{11734B} yes}} & \multicolumn{1}{c|}{{\color[HTML]{B10202} no}}  & \multicolumn{1}{c|}{{\color[HTML]{B10202} no}} & {\color[HTML]{B10202} no}  & \multicolumn{1}{c|}{{\color[HTML]{11734B} yes}} & \multicolumn{1}{c|}{{\color[HTML]{11734B} yes}} & \multicolumn{1}{c|}{{\color[HTML]{B10202} no}}  & {\color[HTML]{B10202} no}  & \multicolumn{1}{c|}{{\color[HTML]{11734B} yes}} & \multicolumn{1}{c|}{{\color[HTML]{11734B} yes}} & \multicolumn{1}{c|}{{\color[HTML]{B10202} no}}  & {\color[HTML]{B10202} no}  \\ \hline
Aggregated Pass Rate & \textbf{92\%} & \textbf{58\%} & \textbf{33\%} & \textbf{25\%} & 75\% & 8\% & 0\% & 0\% & \textbf{92\%} & 42\% &  8\% & 0\% & \textbf{92\%} & 42\% & 8\% & 0\% \\
\bottomrule
\end{tabular}
}
\vspace{3mm}
\caption{\textbf{Additional Experimental Results}: FactorSim outperforms other Chain of Thought baselines by a large margin on assembly tasks. \textit{Syntax} indicates the task passes the syntax check. \textit{Runtime} indicates that the task can run in the physics simulator. \textit{Task} indicates whether the task can  be completed by the oracle agent. \textit{Human} indicates whether the completed task matches the input prompt specification.}
\end{table}

%\begin{figure*}[ht]
%    \centering
%    \includegraphics[width=\textwidth]{figures/robotics_workflow.png}
%    \caption{The workflow of \method on generating robotic tasks.}
%    \label{fig:robotics_workflow}
%\end{figure*}

\begin{lstlisting}[label={list:supp:prompts:gensim1},language={},caption=The chain of thought prompt.]
The game designer is building a single-player game in Pygame by modeling this game as a Markov Decision Process (MDP). Your task is to detect key/mouse input and update the state variables accordingly according to a feature requested by the game designer.
The game has the following implementation already:
```python
import pygame
import sys
import random


{state_manager_code}

# existing input event handling functions
{existing_implementation}
# the new logic function will be here
# if the function is already implemented, it will be replaced with the new implementation

def main():
    state_manager = StateManager()
    running = True
    while running:
        event = pygame.event.poll()
        if event.type == pygame.QUIT:
            running = False
        # {{function_description}}
        {{function_name}}(state_manager, event)
    pygame.quit()

if __name__ == "__main__":
    pygame.init()
    main()
```

To ensure the implementation is correct, please also implmeent an unit test for the function. Please implement the following request from the game designer and return your answer in the following format:
```json
{{
    "function_name": "{function_name}",
    "function_description": "{function_description}",
    "function_implementation": "the pygame implementation of the function, including the first line of the function definition",
    "unit_test": "the unit test code for the function"
}}
```

Here are the dos and don'ts for this request:
- Note that the implementation of the function shuold only have two arguments (i.e. state_manager and event).
- The function implementation should involve checking user input with event (i.e. event.type and event.key).
- Minimize the number of functions added while meeting the game designer's requirements. However, make sure to always give the full implementation of the function.
- Include only the essential details requested by the game designer. Do not add things that are not requested.
- Please use the state variables defined in the state manager. Do not introduce new state variables.
- Only KEYDOWN events will be detected. Do not rely on KEYUP events.
- Check for index out of bounds errors with any lists or arrays being used.
- Check for divide-by-zero errors.
- Do not leave any code incomplete. Do not leave placeholder values. Do not provide demonstration code implementation. Be sure all code is fully implemented.
\end{lstlisting}

\begin{lstlisting}[label={list:supp:prompts:gensim2},language={},caption=The state change prompt.]
**Objective:** Create a Python program that generates state variables for a robot simulation task. The state variables should be able to depict the final target environment for this task.

**Task Details:**
- Task: `TARGET_TASK_NAME`
- Goal: `TARGET_TASK_DESCRIPTION`
- URDFs: `TASK_ASSET_PROMPT`

**Requirements:**
- Return a python function called `get_state_variables` which returns a list of `StateVariableData` objects. We should be able to take the result of this function and plug them into a robotics environment, and it should display what the final target environment will look like for the given task.
- The `StateVariableData` dataclass is defined as follows:
```
@dataclass
class StateVariableData:
    variable_description: str
    variable_name: str
    variable_urdf: str
    variable_size: Tuple[float, float, float]
    variable_color: str
    variable_target_pose: List[Tuple[Tuple[float, float, float], Tuple[float, float, float]]]
    variable_amount: str
    static: bool
```

**Important Notes**
- Item sizes and positions must reflect a realistic setting. Ensure logical spatial relationships; items should neither overlap nor float unnaturally. Account for the items' dimensions and sizes when determining their placement. 
- "variable_name" is a unique name for every state variable.
- "variable_size" should be a list of 3 floats, representing the size of the object along the three dimension [x, y, z].
- "variable_target_pose" should be a list of poses, where each pose is a 2-element list containining the 3-element position vector, and 4-element quaternion rotation. The position vectors should be within the following boundaries: [0.25, 0.75] for the x-axis, [-0.5, 0.5] for the y-axis, and [0.01, 0.3] for the z-axis. For a single object, the target pose should only contain one element, like [[[0.52, 0.02, 0.001], [0, 0, 0, 1]]]. For multiple objects, the target pose can contain multiple items, like [[[0.52, 0.02, 0.001], [0, 0, 0, 1]], [[0.48, 0.02, 0.001], [0, 0, 0, 1]]].
- "static" describes if an object is something part of the environment and is not something the robotic agent should be moving. For example, if the task is to put a ball into the container, the container should be static.

**Example Input and Expected Output: **
- Task: `build_a_car`
- Goal: "Construct a simple car structure using blocks and cylinders."
- URDFs: ["box/box-template.urdf", "cylinder/cylinder-template.urdf"]
- Expected output
```python
def get_state_variables() -> List[StateVariableData]:
    car_pose = ((0.5, 0.0, 0.0), (0,0,0,1))  # fixed pose
    base_length = 0.04

    base_target_pose = [(utils.apply(car_pose, (base_length / 2,  base_length / 2, 0.001)), car_pose[1]),
                    (utils.apply(car_pose, (-base_length / 2,  base_length / 2, 0.001)), car_pose[1])]

    wheel_length = 0.12
    wheel_target_poses = [(utils.apply(car_pose, ( wheel_length / 2,  wheel_length / 2, 0.001)), car_pose[1]),
                            (utils.apply(car_pose, (-wheel_length / 2,  wheel_length / 2, 0.001)), car_pose[1]),
                            (utils.apply(car_pose, ( wheel_length / 2, -wheel_length / 2, 0.001)), car_pose[1]),
                            (utils.apply(car_pose, (-wheel_length / 2, -wheel_length / 2, 0.001)), car_pose[1])]

    return [
        StateVariableData(
            variable_description="blocks used to build the base",
            variable_name="base",
            variable_size=[0.02, 0.04, 0.02],
            variable_urdf="box/box-template.urdf",
            variable_color="red",
            variable_target_pose=base_target_pose,
            variable_amount=2,
            stationary=False,
        ),
        StateVariableData(
            variable_description="wheel to put on the base",
            variable_name="wheel",
            variable_size=[0.02, 0.02, 0.02],
            variable_urdf="cylinder/cylinder-template.urdf",
            variable_color="black",
            variable_target_pose=wheel_target_poses,
            variable_amount=4
            stationary=False,
        ),
        StateVariableData(
            variable_description="body of the car",
            variable_name="body",
            variable_size=[0.04, 0.02, 0.02],
            variable_urdf="box/box-template.urdf",
            variable_color="blue",
            variable_target_pose":[car_pose],
            variable_amount=1
            stationary=False,
        )
    ]
```

**Example Input and Expected Output: **
- Task: `build_a_circle`
- Goal: "Construct a circle using 6 red blocks"
- URDFs: ["block/block.urdf"]
- Expected output
```python
def get_state_variables() -> List[StateVariableData]:
    block_size = (0.04, 0.04, 0.04)

    red_circle_poses = []
    circle_radius = 0.1
    circle_center = (0, 0, block_size[2] / 2)
    angles = np.linspace(0, 2 * np.pi, 6, endpoint=False)
    circle_pose = ((0.4, 0.3, 0.0), (0, 0, 0, 1))  # fixed pose

    # Define initial and target poses for the red and blue circles.
    for angle in angles:
        pos =  (circle_center[0] + circle_radius * np.cos(angle),
                circle_center[1] + circle_radius * np.sin(angle),
                circle_center[2])
        block_pose = (utils.apply(circle_pose, pos), circle_pose[1])
        red_circle_poses.append(block_pose)


    return [
        StateVariableData(
            variable_description="blocks used to build the base",
            variable_name="base",
            variable_size=block_size,
            variable_urdf="block/block.urdf",
            variable_color="red",
            variable_target_pose=red_circle_poses,
            variable_amount=10,
            stationary=False,
        ),
    ]
```
\end{lstlisting}

\begin{lstlisting}[label={list:supp:prompts:gensim3},language={},caption=
The subtask code generation prompt .]
**Objective:** You are designing a training plan for a robotic arm in a simulation environment to complete a task `TARGET_TASK_NAME`. This task will be completed in multiple subtasks from the subtask list. Each subtask manages the robotic arm to move the composition of assets from one state to another, ultimately achieving the ideal state that completes the task.
For this step, you are asked to generate the subtask function for `SUBTASK_NAME`. It involves `SUBTASK_DESCRIPTION`. Refer to the existing template and use the existing variables to inform your subtask creation. You will generate Python code for {state_variable_for_SUBTASK_NAME} and {subtask_code_for_SUBTASK_NAME}, and return the code with the given JSON format.

**Task Overview:**
- Task Name: `TARGET_TASK_NAME`
- Task Description: `TARGET_TASK_DESCRIPTION`
- All Subtask Descriptions: `TARGET_SUBTASK_LIST`
- Current Subtask Name: `SUBTASK_NAME`
- Current Subtask Description: `SUBTASK_DESCRIPTION`


**Existing template**
SUBTASK_CODE_TEMPLATE


**Subtask Requirements:**
- Generate Python code addressing the specified subtask based on the provided description and initial variables.
- Adhere to guidelines: use specified APIs you just reviewed, avoid unknown libraries, and comment on the code for clarity.
- Ensure all state manager variables used in the subtask code are already defined.


**Subtask function important Notes:**
- Only one `self.add_goal` should be used, and it should not be used on state variables marked with static=True.
- The `matches` argument in the method called `self.add_goal` should always be a numpy array.
- Do not use libraries, functions, and assets that you don't know.
- Do not create additional functions inside the subtask function, only return one function.
- Do not add state variables marked as "static = True" to the environment using `env.add_object`. 
- When passing in color using `env.add_object`, remember to pass it using utils.COLORS[item_color]. Good example is: `env.add_object(base_block_urdf, base_block_pose, color=utils.COLORS['red'])`. Bad example is: `env.add_object(base_block_urdf, base_block_pose, 'red')`
- In the subtask code, you need to create both the initial pose and the target pose. The Initial pose should NOT be very close to or be the same as the target pose. Initial pose determines where the object will be placed before the training starts. Target pose determines the ideal pose where the robot arm will receive reward if placed right in the simulation. You should prioritize using the target_pose from the state manager variables.
- Only use get_random_pose for initial pose(position and rotation). `get_random_pose(env, obj_size)` gets random collision-free object pose within workspace bounds. param obj_size: (3, ) contains the object size in x,y,z dimensions. return: translation (3, ), rotation (4, ). You should pass the obj_size directly to the get_random_pose function, instead of its single element. Good example is: self.get_random_pose(env, block_size). Bad example is: self.get_random_pose(env, block_size[0]) or env.get_random_pose(block_size).
- If you are generating a pose from `get_random_pose`. It will be within bounds. You don't need to check it using other helper function like `is_pose_within_bounds`. Do not use `np.copy` to copy it.
- If you need initial rotation, use `get_random_pose` to get the `pose` first. Then use `pose[1]` as rotation. Do not initiate it in other ways, a bad example is `rotation = np.float32(p.getMatrixFromQuaternion(pose[1])).reshape(3, 3)`.
- DO NOT USE `pose = p.getBasePositionAndOrientation(object_id)`. It's for environment simulation.
- DO NOT USE other unlisted way to create `pose`.
- In self.add_goal, make sure to set "step_max_reward=1./SUBTASK_COUNT"
- Each `env.add_object` call will create a new object id. If only one object is needed in this subtask, then pass in `objs = [object_id]` in `self.add_goal`. If multiple objects are needed, create a list that contains all the needed object ids then pass in the list: `objs = object_id_list`.
- Make sure you include all the arguments to `self.add_goal`: `objs`, `matches`, `targ_poses`, `replace`, `rotations`, `metric`, `params`, `step_max_reward`, `language_goal`.
- For `self.add_goal`'s argument `matches`, it should be `matches=np.ones((n, n))`. `n` represents the total amount of objects added to the `env`. E.g. `env.add_object` was called 4 times then it should be set as `matches=np.ones((4, 4))`.
- You have been given all the task variables for creating the subtask. Do not assume any unknown variables.
- Only three functions are available from `utils`: `utils.apply`, `utils.quatXYZW_to_eulerXYZ` and `utils.COLORS`. Do not make up any other functions from `utils`.
- Do not include triple quotes (""") in your code, only use `#` for comments.
- If the asset of this subtask involves `zone`, make sure that pose of the zone should not be moved, it's usually used for creating target position for other items. E.g:
`zone_size = [0.12, 0.12, 0]
zone_urdf = 'zone/zone.urdf'
zone_colors = ['yellow', 'blue', 'green']
zone_poses = []
for color in zone_colors:
    zone_pose = self.get_random_pose(env, zone_size)
    env.add_object(zone_urdf, zone_pose, 'fixed', color=utils.COLORS[color])
    zone_poses.append(zone_pose)`
- If the subtask involves updating the rotation, you may call `utils.quatXYZW_to_eulerXYZ`. Here's what this function do:
`def quatXYZW_to_eulerXYZ(quaternion_xyzw):  # pylint: disable=invalid-name
    """Abstraction for converting from quaternion to a 3-parameter rotation.

    This will help us switch which rotation parameterization we use.
    Quaternion should be in xyzw order for pybullet.

    Args:
      quaternion_xyzw: in xyzw order, tuple of 4 floats

    Returns:
      rotation: a 3-parameter rotation, in xyz order, tuple of 3 floats
    """
    q = quaternion_xyzw
    quaternion_wxyz = np.array([q[3], q[0], q[1], q[2]])
    euler_zxy = euler.quat2euler(quaternion_wxyz, axes='szxy')
    euler_xyz = (euler_zxy[1], euler_zxy[2], euler_zxy[0])
    return euler_xyz
`

**Example Input and Expected Output:**
- Task Name: `build_a_car`
- Task Description: `Construct a simple car structure using blocks and cylinders.`
- Subtask List: `["build_car_base: Build the base of the car in the simulation environment.", "build_car_body: Build the body of the car in the simulation environment.", "build_car_wheels: Build the wheels of the car in the simulation environment."]`
- Subtask Name: `build_car_base`
- Subtask Description: `Build the base of the car in the simulation environment.`
- Existing template:

import numpy as np
from cliport.tasks.task import Task
from cliport.utils import utils


class MyBuildCar(Task):
    """Construct a simple car structure using blocks and cylinders."""

    def __init__(self):
        super().__init__()
        # initialize the state_manager attributes that manage all the state variables in this task
        StateManager = type('StateManager', (object,), {})
        self.state_manager = StateManager()
        self.state_manager.main_target_pose = [[0.5, 0.0, 0.0], [0, 0, 0, 1]]
        self.max_steps = 15
        self.state_manager.base_size = (0.04, 0.08, 0.02)
        self.state_manager.base_color = "green"
        self.state_manager.base_urdf = "box/box-template.urdf"
        self.state_manager.base_amount = 2
        self.state_manager.anchor_base_poses = [((0.52, 0.02, 0.001), (0, 0, 0, 1)), ((0.48, 0.02, 0.001), (0, 0, 0, 1))]
        self.additional_reset()

{placeholder_subtask_code}

    def reset(self, env):
        super().reset(env)
        self.place_first_step(env)


- Expected output:
```python
def build_car_base(self, env):
    # Build the base of the car in the simulation environment. Let's solve this problem step-by-step.

    # Step 1. Retrieve the variables(the car's pose, the base length, and the base size) to initialize the car base building.
    base_size = self.state_manager.base_size
    base_color = self.state_manager.base_color
    base_amount = self.state_manager.base_amount
    anchor_base_poses = self.state_manager.anchor_base_poses

    # Step 2. Setting up the base block URDF.
    base_urdf_path = self.state_manager.base_urdf
    base_block_urdf = self.fill_template(base_block_urdf,  {'DIM': base_size})

    # Step 3: Adding base blocks to the scene
    base_blocks = []
    for idx in range(2):
        base_block_pose = self.get_random_pose(env, base_size)
        base_block_id = env.add_object(base_block_urdf, base_block_pose, color=utils.COLORS['red'])
        base_blocks.append(base_block_id)

    # Step 4: Setting the goal to create the base of the car by positioning two red blocks side by side.
    self.add_goal(
        objs=base_blocks,
        matches=np.ones((base_amount, base_amount)),
        targ_poses=anchor_base_poses,
        replace=False,
        rotations=True,
        metric='pose',
        params=None,
        step_max_reward=1./3,
        language_goal="Firstly, create the base of the car by positioning two red blocks side by side."
    )
```

REFLECT_PROMPT

**Output Format:**
- Use `SUBTASK_NAME` as the function name.
- IMPORTANT doublecheck your code is following everything in important notes.
\end{lstlisting}

\section{Additional details for the proposed generative simulation benchmark}
This section provides the prompts for all 8 RL games in the benchmark.
\begin{lstlisting}[label={list:supp:prompts:catcher},language={},caption=
The prompt for the game \textit{Catcher}.]
Create a catcher character, represented as a rectangle, positioned at the bottom and the middle of the screen.
Allow the player to control the catcher's horizontal movement using the left and right arrow keys on the keyboard.
There should always be exactly one ball on the screen at all times. The ball should be visually distinct and easily recognizable.
Make the ball move downwards at a steady pace towards the catcher. The speed can be constant or increase gradually as the game progresses.
Detect collisions between the catcher and the ball. When the catcher catches a ball, increment the player's score, spawn a new ball, and display this score in the top-left corner of the screen.
Give the player a 3 lives. Each time a ball is missed by the catcher and reaches the bottom of the screen, decrease the player's life count by one.
End the game when the player's lives reach zero. Display a "Game Over!" message and temporarily halt gameplay but dont terminate the game.
Provide an option for the player to click the screen to restart the game after the "Game Over" screen is displayed.
Continuously generate new balls after each catch or miss, ensuring endless gameplay. Optionally, increase the game's difficulty gradually by speeding up the ball's fall or reducing the size of the catcher as the player's score increases.
\end{lstlisting}
\begin{lstlisting}[label={list:supp:prompts:flappy_bird},language={},caption=
The prompt for the game \textit{Flappy Bird}.]
Create a bird character, visually represented as a simple rectangle within the game window.
Introduce gravity, causing the bird to continuously fall slowly.
Allow the bird to 'jump' or accelerate upwards in response to a player's mouse click, temporarily overcoming gravity.
Periodically spawn pairs of vertical pipes moving from right to left across the screen. Each pair should have a gap for the bird to pass through, and their heights should vary randomly.
If the bird makes contact with the ground, pipes or goes above the top of the screen the game is over.
Implement the following scoring system: for each pipe it passes through it gains a positive reward of +1. Each time a terminal state is reached it receives a negative reward of -1.
When the game ends, display a "Game Over!" messagea and stop all the motion of the game.
Show the current score in the top-left corner of the screen during gameplay.
Ensure the game has no predefined end and that new pipes continue to generate, maintaining consistent difficulty as the game progresses.
\end{lstlisting}
\begin{lstlisting}[label={list:supp:prompts:snake},language={},caption=
The prompt for the game \textit{Snake}.]
Create a snake character represented as a series of connected pixels or blocks. Initially, the snake should be a single block (i.e. the head) that moves in a specific direction within the game window.
Allow the player to control the snake's movement using arrow keys. The snake should be able to turn left or right, but it should not be able to move directly backward. Eg. if its moving downwards it cannot move up.
The movement of the snake should be continuous in the current direction until the player provides new input. Ensure that the snake moves one grid unit at a time.
Implement a basic food system where one food item appears randomly on the screen.
When the snake consumes the food by moving over or colliding with it, the snake's length increases, and the player earns points. It recieves a positive reward, +1, for each food the head comes in contact with. While getting -1 for each terminal state it reaches.
If the head of the snake comes in contact with any of the walls or its own body, the game should end.
Incorporate a scoring system, displaying the current score on the screen during gameplay. The score should increase each time the snake consumes food.
Ensure that the game has no predefined end, allowing the snake to continue growing and the difficulty to increase over time. New food items should appear after the snake consumes one.
Provide an option for the player to restart the game after it ends. Display a "Restart" option on the game over screen to allow the player to play again.
\end{lstlisting}
\begin{lstlisting}[label={list:supp:prompts:pixelcopter},language={},caption=
The prompt for the game \textit{Pixelcopter}.]
Create a copter character represented as a large white square that remains fixed horizontally but can ascend and descend vertically within the game window.
Introduce gravity mechanics, causing the copter to continuously descend slowly. Enable the player to ascend when the player clicks the mouse, allowing it to momentarily counteract gravity and rise upwards.
Create obstacles in the shape of a cavern. Construct the cavern using a series of vertically aligned rectangular barriers positioned at both the bottom and the top of the screen. Ensure the adjacent obstacles are of similar length to maintain a consistently smooth "tunnel" effect.
Implement collision detection to detect when the copter collides with obstacles or the boundaries of the game window, triggering the end of the game upon collision.
Display a "Game Over!" message prominently when the game ends due to a collision, halting all movement within the game and prompting the player to restart.
Create a scoring system that rewards the player based on how far the copter travels through the maze without colliding with obstacles.
Show the current score in the top left area of the screen.
Ensure the game has no predefined end and that new obstacles continue to generate, maintaining consistent difficulty as the game progresses
Allow the player to start a new game after a collision.
\end{lstlisting}
\begin{lstlisting}[label={list:supp:prompts:pong},language={},caption=
The prompt for the game \textit{Pong}.]
Create a paddle character for the human player, represented as a rectangle, positioned on the left side of the screen. 
Allow the human player to control the paddle's vertical movement using the up and down arrow keys. The paddle has a little velocity added to it to allow smooth movements.
Implement a paddle character for the CPU opponent, also represented as a rectangle, positioned on the opposite side of the screen.
Introduce a ball that moves across the screen with a speed. The ball should bounce off the paddles and the top and bottom walls of the game window.
If the ball goes off the left or right side of the screen, reset its position to the center and its direction.
The CPU to control its paddle's vertical movement to autonomously track the ball.
Detect collisions between the ball and the paddles. When the ball collides with a paddle, make it bounce off in the opposite direction.
Implement a scoring system that rewards the human player for returning the ball back (i.e. making contact with the bal).
Display the current score at the top of the screen. Ensure the game has no predefined end, allowing for continuous play. 
Display a "Game Over!" message when one player reaches a score of 5 and provide an option for the player to restart the game after the "Game Over" screen is displayed.
\end{lstlisting}
\begin{lstlisting}[label={list:supp:prompts:puckworld},language={},caption=
The prompt for the game \textit{Puckworld}.]
Create an agent character, visually represented as a blue circle, positioned on the screen. The agent should be movable in any direction based on user input.
Implement a green dot that moves randomly around the screen, serving as the target for the agent to navigate towards.
Introduce a red puck, a larger entity that slowly follows the agent's movements on the screen.
Allow the player to control the agent's movement using arrow keys or another specified input method.
Implement a scoring system that positively rewards the agent proportionally to the closeness between the agent and the green dot, and penalizes the agent for proximity to the red puck.
Display the current score in the top-left corner of the screen during gameplay.
Ensure the game has no predefined end, allowing for endless gameplay. 
Upon reaching the green dot, relocate it to a new random position, maintaining the challenge for the player.
\end{lstlisting}
\begin{lstlisting}[label={list:supp:prompts:waterworld},language={},caption=
The prompt for the game \textit{Waterworld}.]
Create a player character visually represented as a blue circle that can move freely within the game window using arrow keys.
Introduce a dynamic environment with equal number of green and red circles. Make the green and red circles move randomly around the screen.
Implement a scoring system, where the player earns points for each green circle captured and deduct one for each red circle . 
Display the current score in the top-left corner of the screen during gameplay.
When the player captures a circle, make it respawn in a random location on the screen as either red or green.
Ensure the game continues until all green circles have been captured. Once all green circles are captured, display a "Game Over!" message and stop all motion in the game.
Provide an option for the player to restart the game after it ends, creating a loop for continuous gameplay.
\end{lstlisting}
\begin{lstlisting}[label={list:supp:prompts:monster_kong},language={},caption=
The prompt for the game \textit{Monster Kong}.]
Write code for a 2d game viewed from the side where the character can walk and jump. Let the character move left or right using the 'a' and 'd' keys. Let the character jump using the 'w' key.
Create a level by arranging 5 stationary platforms above the ground. Make sure the character's jump can reach the platform height.
Let the character stand on the ground or platforms but fall otherwise. Start the player on the ground.
Add a princess character that the character must rescue by reaching her position. Place the princess on one of the platforms.
Implement fireballs that fall from random places from the top of the screen. Do not let the fireballs move through the platforms. These fireballs serve as obstacles that the player must avoid.
Touching a fireball should deduct 25 points from the player's score and cause them to lose a life. The game ends if the player loses fifteen lives.
Scatter ten coins randomly around the game window that the player can collect for 5 points each.
Award the player 50 points for rescuing the princess. Move the princess to a random platform when the player rescues her.
Display the current score and remaining lives in the game window.
\end{lstlisting}

\section{Additional details for the human study experiment}
\begin{figure}
  \centering
   \caption{Human study results on all 8 games.}
       \vspace{2em}
  \includegraphics[width=\linewidth]{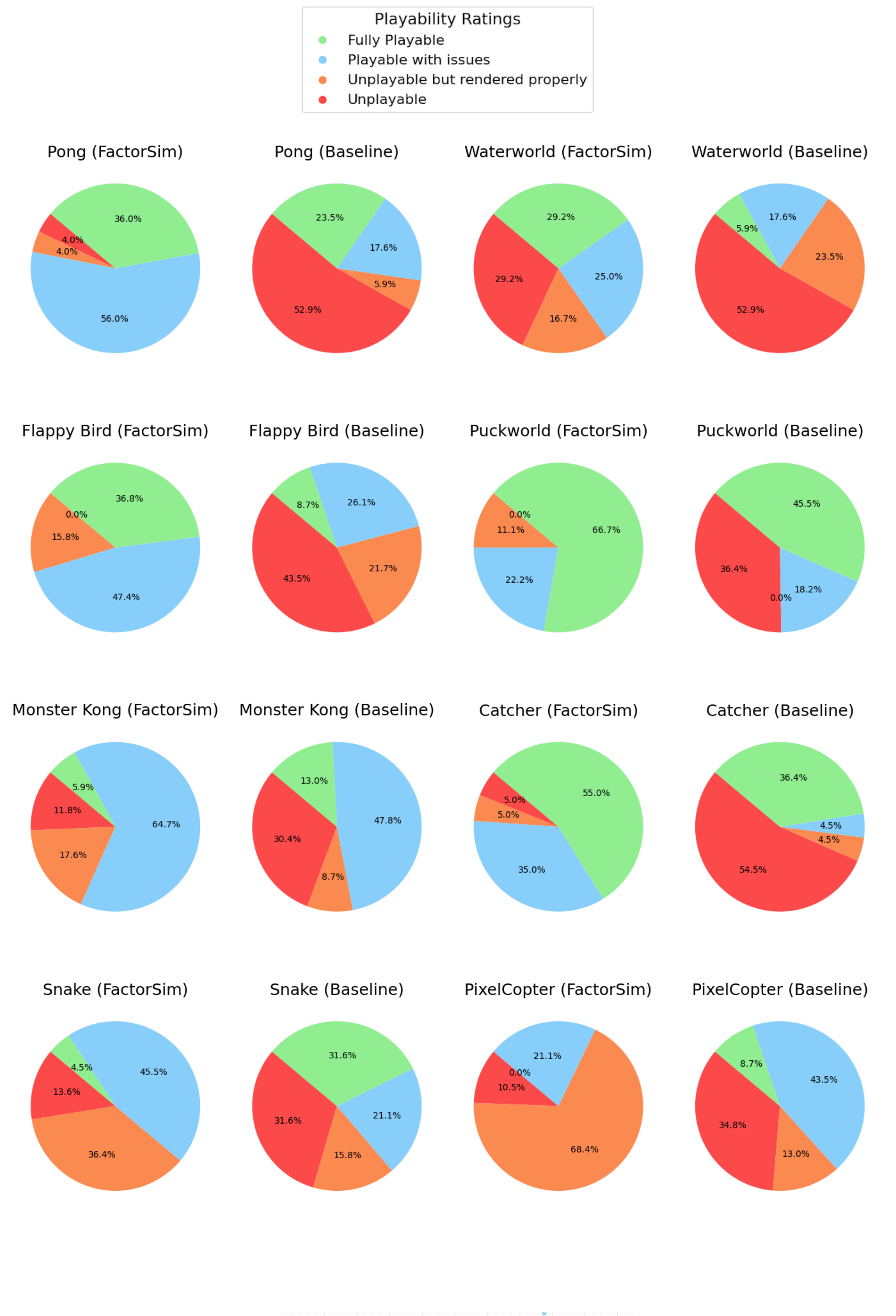}
  \label{fig:all_human_study_result}
\end{figure}

In this section, we first provide details for the experiment and then the instructions we gave to human participants in our human study, along with the user interface. We also provide the detailed results of this evaluation for all games in Figure~\ref{fig:all_human_study_result}.

Human participants were asked to play and evaluate the generated games given the prompt while excluding factors such as aesthetics or difficulty. They rated the games on a scale of 1 to 4, where 4 indicates a fully playable game, 3 is a playable game with some bugs or flaws that hinder gameplay experience, 2 is an unplayable game (i.e., no interactivity) with correctly rendered UI, and 1 is a game that crashes or fails to launch.

\begin{lstlisting}[label={list:supp:prompts:human study},language={},caption=The instructions we give to human participants to our human study.]
Welcome to your user study! Your task is to evaluate AI-generated games.
Select the game you want to generate and click the button "Generate" to generate games.
You might have to wait for the game to load for 5-10 seconds.

Note that the game is intentionally slowed down, making it easier for you to evaluate them!
Thus, when you click or press a key/button, the "character" might react slower than you expected.
Please click "Random Generate" to generate a game.

- If the game doesn't load (black screen), select "1 - unplayable."
- Please do not consider the difficulty of the game.
- Please don't take aesthetics into account.
- You want to assess whether the UI elements are rendered accurately for gameplay purposes while excluding considerations related to aesthetics, overlapping, or duplicated UI components.

Given the prompt as shown, please judge the playability of the game.
- 4: Fully Playable: the game is generally playable from start to finish without significant bugs.
- 3: Playable but with some flaws: the game is somewhat playable (interactable), but there are issues (inaccurate logic or glitches) that impair the gameplay.
- 2: Not playable: no interactivity but the UI seems to be rendered correctly.
- 1: Unplayable: the game cannot be started, or it crashes immediately upon launch.
  
\end{lstlisting}

\begin{figure}
  \centering
  \includegraphics[width=\linewidth]{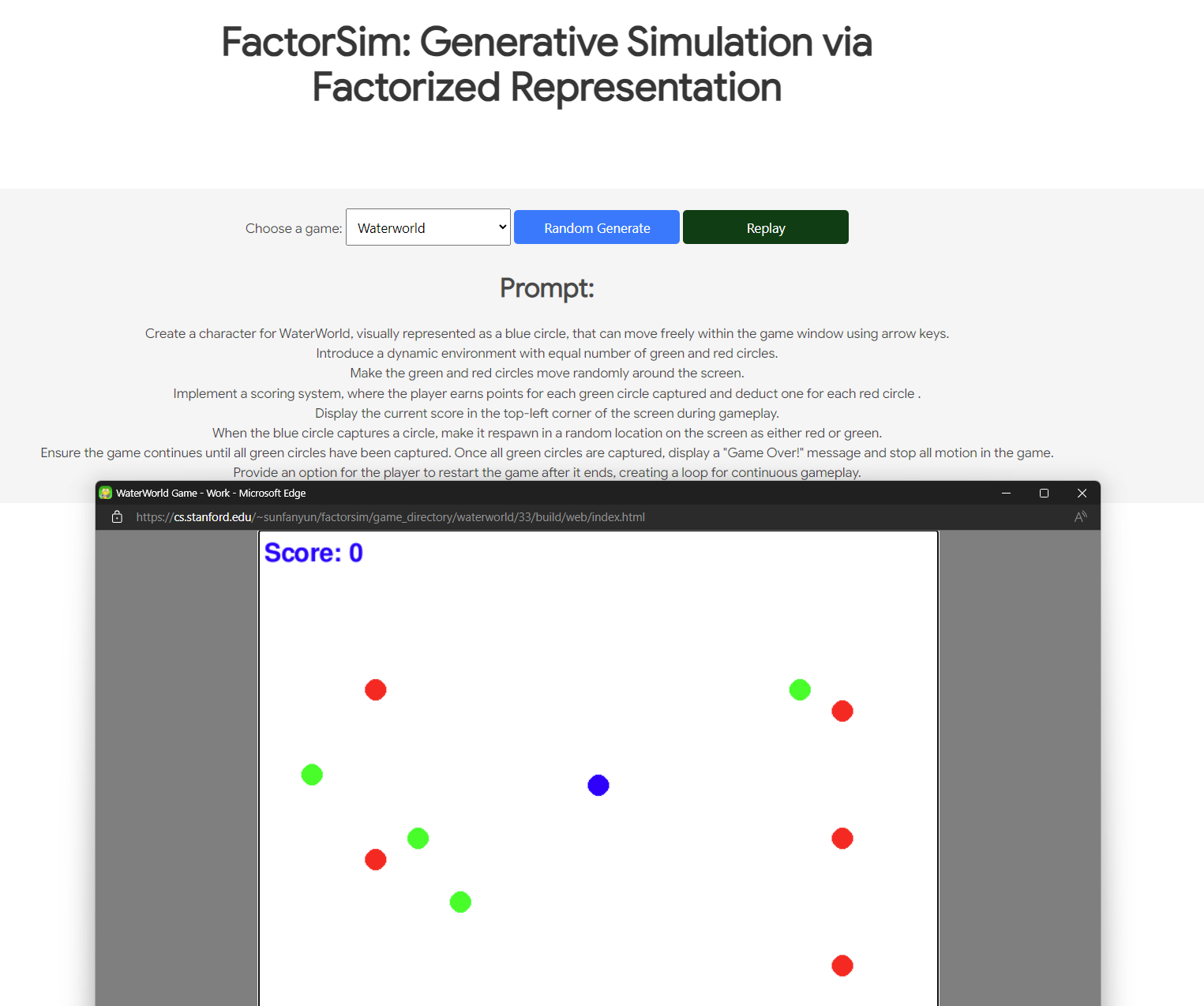}
   \caption{Human study interface screenshot.}
  \label{fig:human_study_interface}
\end{figure}

%\clearpage
%\pagebreak
%\input{checklist}

\end{document}